\pdfoutput=1
\documentclass[11pt]{article}

\usepackage[final]{acl}
\usepackage[most]{tcolorbox}
\usepackage{times}
\usepackage{latexsym}
\usepackage{adjustbox}
\usepackage{enumitem}
\usepackage[T1]{fontenc}

\usepackage[utf8]{inputenc}

\usepackage{microtype}

\usepackage{inconsolata}

\usepackage{graphicx}
\usepackage{tabularx}
\usepackage{float} % For [H]

\usepackage{algorithm}
\usepackage{algorithmic}

\usepackage{amsmath}
\usepackage{amssymb}
\usepackage{booktabs}
\usepackage{caption}
\usepackage{subcaption}
\usepackage{appendix}
\usepackage{makecell}
\usepackage{multirow}

\newcommand{\narmeen}[1]{\textcolor{blue}{[Narmeen: #1]}}

\title{K-Steering: Multi-Dimensional Control of Large Language Models Using Non-Linear Classifier}
\title{Beyond Linear Steering: Unified Multi-Attribute Control for Language Models}

\author{
Narmeen Oozeer\thanks{Equal contribution}\thanks{Corresponding authors: \texttt{narmeen@withmartian.com, amir.abdullah@thoughtworks.com}}\\
\textit{Martian}
\And
Luke Marks\footnotemark[1]\\
\textit{Martian}
\And
Shreyans Jain\\
\textit{Martian}\\
\And
Fazl Barez\\
\textit{University of Oxford}\\
\textit{Whitebox}
\And
Amirali Abdullah \footnotemark[2]\thanks{Advising author}\\
\textit{Thoughtworks} \\
\textit{Martian}
}

\begin{document}
\maketitle

\begin{abstract}

\iffalse
Inference-time interventions on token hidden states (aka ``steering'') enable attribute control over LLM generations. Existing multi-attribute methods typically sum separate steering vectors, incurring performance trade-offs when attributes interact, as well as significant overhead of sweeping over hyperparameters.  We introduce K-Steering, a unified approach that in contrast learns a \textit{single} steering vector via the gradients of a non-linear classifier, adapting to attribute boundaries in activation space. Experiments across two complex datasets and 3 model families show K-Steering outperforms baseline methods.
\fi

Controlling multiple behavioral attributes in large language models (LLMs) at inference time is a challenging problem due to interference between attributes and the limitations of linear steering methods, which assume additive behavior in activation space and require per-attribute tuning. We introduce K-Steering, a unified and flexible approach that trains a single non-linear multi-label classifier on hidden activations and computes intervention directions via gradients at inference time. This avoids linearity assumptions, removes the need for storing and tuning separate attribute vectors, and allows dynamic composition of behaviors without retraining. To evaluate our method, we propose two new benchmarks, \textsc{ToneBank} and \textsc{DebateMix}, targeting compositional behavioral control. Empirical results across 3 model families, validated by both activation-based classifiers and LLM-based judges, demonstrate that K-Steering outperforms strong baselines in accurately steering multiple behaviors.

The implementation of the K-steering framework is available in our public repository 
\footnote{
Code available at: \url{https://github.com/withmartian/k-steering}
}

\end{abstract}

\section{Introduction}
Large language models (LLMs) are increasingly deployed in high-stakes settings, necessitating reliable mechanisms to guide their behavior at inference time. While alignment methods such as Reinforcement Learning from Human Feedback (RLHF) \citep{ouyangrlhf} have improved overall model helpfulness and harmlessness, they are expensive and inflexible to deploy \citep{expensive_continual}, may inadvertently make models \textit{less} safe \citep{bad_finetune} and cause model forgetting \citep{forgetting}. In contrast, activation-based steering methods offer lightweight alternatives that modify internal representations at inference time to influence attributes such as tone, factuality, or toxicity without retraining the full model \citep{wu2025axbench,dathathri2019plug, peft_alignment}.

%\fazl{i feel like in here the best comparison would be finetuning methods like lora} \amir{Added some citations using peft}

Most existing approaches focus on single-attribute control, typically using linear interventions. These include difference-in-classes methods \cite{bolukbasi, phillips_means, marks2023geometry, arditi2024refusal}, vector-based edits \cite{meng2022mass}, and gradient-guided control using small auxiliary classifiers \citep{dathathri2019plug}. Extending such techniques to multi-attribute control is non-trivial; for instance, a naive extension by simply combining steering vectors often leads to interference or degraded fluency \citep{adding_multiple, nguyen2025multi, song2025effectively}, as well as separate hyperparameter sweeps required for computing each individual vector. More recent frameworks like MAT-STEER \citep{nguyen2025multi} and CONFST 
\citep{song2025effectively} attempt to address this through orthogonality constraints and confident vector selection. However, they still assume an additive linear structure of the steering vectors and require per-attribute gating functions to be learnt on the activations at each token.

\iffalse
In this paper, we introduce K-Steering, a gradient-based, classifier-driven method for multi-dimensional control of LLM outputs. Rather than computing separate steering vectors for each attribute, K-Steering trains a single non-linear classifier over the model's activation space to predict $K$ behavioral attributes. At inference time, the classifier’s gradients guide minimal, context-aware interventions that simultaneously steer generation toward (or away from) a specified set of attributes. This avoids the ``watering down" effects of linearly combining vectors \citep{adding_multiple} and supports compositional control across domains such as conversational tones, debate styles and AI safety desiderata such as toxicity mitigation, truthfulness and debiasing.
\fi

In this paper, we introduce \textit{K-Steering}, a method that extends non-linear gradient-based steering approaches \citep{casper_nonlinear} from single-attribute control to multi-attribute control settings. K-Steering fills an important gap in current methods by enabling simultaneous control over multiple behaviors using a single, unified classifier. Unlike prior approaches that steer each attribute independently, K-Steering learns a joint representation of multiple behaviors by training a classifier on hidden activations. At inference time, it uses the classifier’s gradients to adjust the model’s behavior toward (or away from) a specified set of target attributes. This approach avoids the assumption that behaviors compose linearly, handles inter-attribute interactions more robustly, and scales naturally to larger attribute sets.

Empirically, we find that an activation classifier trained at the final layer detects a stronger presence of target class behaviors in K-Steered outputs compared to baselines that average independent steering vectors. These results are supported by traditional steering evaluation methods, such as LLM judges, which indicate that K-Steering better resists the dilution effect of averaging multiple vectors.

Our contributions are three-fold:

\begin{enumerate}
    \item \textbf{Datasets}: We introduce two novel datasets specifically designed for multi-attribute steering with K-Steering: \textsc{TONEBANK} for controlling conversational tones and \textsc{DEBATEMIX} for steering across different debate styles.
    
    \item \textbf{Methodology}: We demonstrate that K-Steering effectively steers across multiple directions simultaneously, performing on par with or better than established baselines. By using additional gradient steps with smaller step sizes, we can smoothly improve steering performance. However, for larger step sizes, multi-step interventions can lead to degraded outputs. We also show that K-Steering directions can be ablated to remove behaviors, analogous to difference-in-means vector ablation as used by \citet{arditi2024refusal}.
    
    \item \textbf{Evaluation Framework}: We develop a semi-automated evaluation protocol that calibrates steering magnitudes for both K-Steering and baseline methods. This protocol uses binary search to find the optimal magnitudes that maximize steering effectiveness while minimizing distribution shift.
\end{enumerate}

The remainder of the paper is structured as follows. Section \ref{sec:related-work} gives an overview of existing steering techniques and their shortcomings. Section \ref{sec:methoddology} introduces our proposed method, K-Steering, including its formalization and gradient-based intervention algorithms. Section \ref{sec:datasets} describes the datasets and model configurations used in our experiments, with dataset construction details and representative examples provided in Appendices \ref{sec:dataset_creation} and \ref{sec:dataset_examples}, respectively. We also give a list of the attributes we use as steering targets for each dataset in Appendix \ref{sec:labels}. In Section \ref{sec:single-step-single-layer}, we compare K-Steering to baselines including CAA and DCT; details of the LLM-based evaluation protocol are provided in Appendix \ref{sec:coherence_judge} and Appendix \ref{sec:steering_judge}. Section \ref{sec:in-depth-analysis} analyzes K-Steering hyperparameters, including multi-layer and multi-step steering, with computational cost estimates included in Appendix \ref{sec:cost_comparison}. Section \ref{sec:projection-removal-results} evaluates projection removal (Algorithm \ref{alg:projection-removal}) as an efficient, one-step ablation method. Appendix \ref{sec:k-steered-completions} contains representative samples of K-Steered generation examples. Section \ref{sec:conclusion} concludes with broader implications and future directions.

\iffalse
Rather than computing separate steering vectors for each attribute, K-Steering trains a single non-linear classifier over the model's activation space to predict $K$ behavioral attributes. At inference time, the classifier's gradients guide minimal, context-aware interventions that simultaneously steer generation toward (or away from) a specified set of attributes. This can avoid the ``watering down'' effects of linearly combining vectors and supports better compositional control of multiple steering attributes.
\fi
\section{Background and Related Work}
\label{sec:related-work}
\subsection{Language Model Alignment}
Language model alignment refers to the process of conforming model outputs with human preferences and values. Current alignment techniques include Reinforcement Learning from Human Feedback (RLHF) \citep{ouyangrlhf, rlhf_2, rlhf_3, rlhf_4}, Constitutional AI \citep{constitutional_1, constitutional_2}, and various forms of instruction tuning. While these methods have been successful in improving model behavior, they typically operate on a single dimension (e.g., helpfulness or harmlessness), and may encode alignment constraints in ways that are vulnerable to manipulation.

\subsection{Activation Steering and Editing}
Recent work has explored manipulating language model behavior by directly intervening in the model's activation space. \citet{zou2023representation}, \cite{inference_li}, and \citet{turner2023activation} demonstrated that by identifying directions in the activation space corresponding to specific attributes, one can steer model generations by adding or removing projections along these directions. However, the most popular methods only focus on single-attribute interventions, and compute a steering vector using first-order, linear methods on hidden states. For example, "inference-time-intervention" (ITI) \citep{inference_li} learns a logistic regression model on the hidden states of positive vs negative examples, while "representation engineering" (RepE) \citep{zou2023representation} extracts the first principal component of the difference between embeddings of positive and negative examples, and contrastive activation addition (CAA) \citep{caa} returns the difference-of-means vector between the hidden states of two classes.

\subsection{Non-linear representations}
Non-linear patterns in activation space have received less attention in the literature, with selected investigations showing progress. \citet{casper_nonlinear} show that interventions guided by non-linear probe gradients are more effective in jailbreaking models, and is the main technical precursor of this work that we extend to handle k attributes at once. \citet{othello_nonlinear} show non-linear probes detect Othello board game states more accurately, and \citet{non_linear_iti} uses non-linear probes to identify attention heads linked with truthful behavior.

\subsection{Multi-dimensional Control}
Controlling language models along multiple dimensions simultaneously has received limited attention. \citet{adding_multiple} show that adding steering vectors corresponding to multiple attributes is ``largely unsuccessful". CONFST \citep{song2025effectively} uses Bayesian learning to identify the most relevant activations from a set of user inputs for an attribute, and then simply averages them to get the steering vector for a single attributes. MAT-STEER \citep{nguyen2025multi} learns a gating function to apply different steering intensities at each token, along with orthogonality constraints to disentangle each steering vector. Our approach is both simpler and more flexible, and uses a single non-linear classifier to model the relationships between multiple tone categories in a shared activation space.

\section{Methodology}
\label{sec:methoddology}
\begin{figure*}[!t]
\centering
\includegraphics[width=0.83\textwidth]{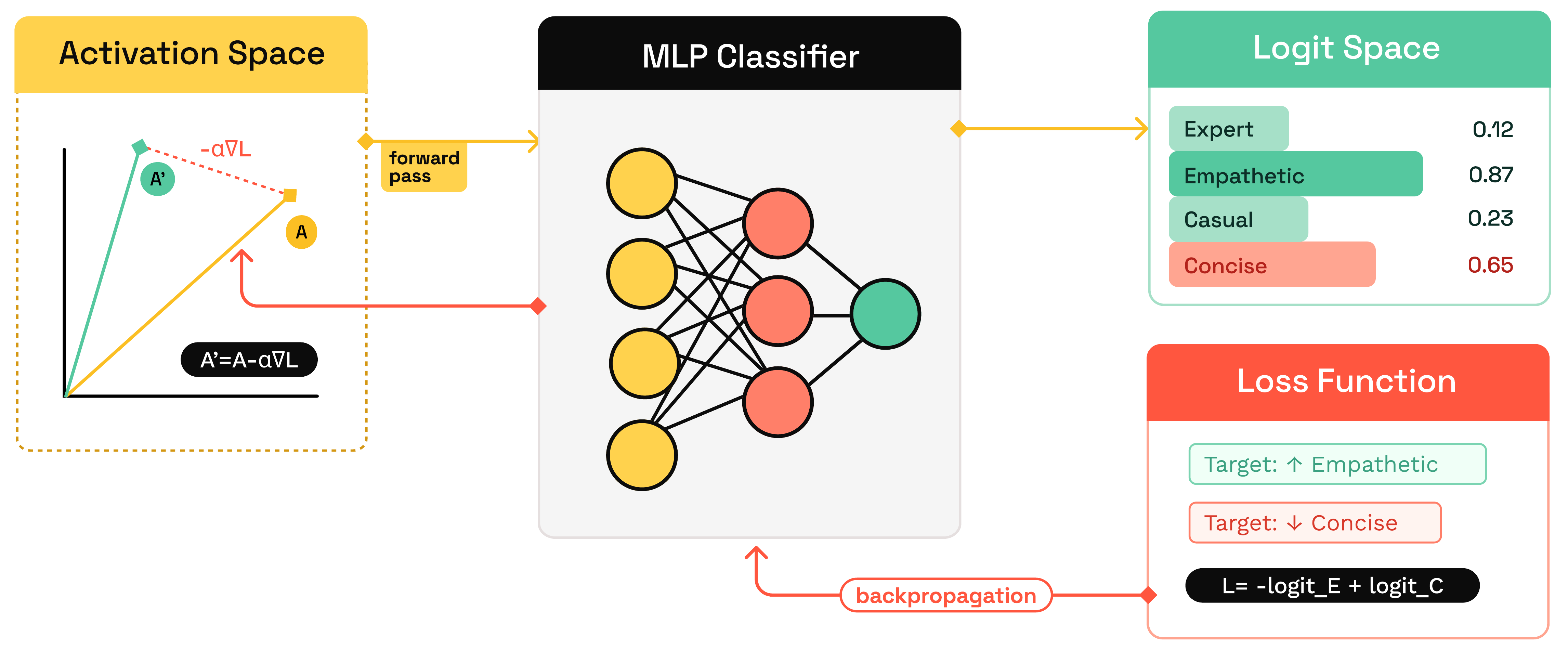}
\caption{An illustration of gradient-based K-Steering. For an activation vector $A$, we calculate a steering loss that penalizes higher logits from a classifier on $A$ for undesired labels and rewards higher logits for desired labels. By backpropagating this loss through the classifier, we obtain the steered activations $A'=A-\alpha\Delta L$.}
\label{fig:your_label}
\end{figure*}

\iffalse
\section{Theoretical Framework}
\subsection{Attributes as Multi-dimensional Constructs}
We conceptualize attributes in language model outputs (tone, factuality, bias, etc.) as multi-dimensional constructs. Different attributes are not simply opposites of each other but exist in complex, non-linear relationships in the model's representation space. Our K-Steering approach aims to capture these relationships and use them to control model outputs across multiple domains simultaneously.

\subsection{Cross-domain Representation Sharing}
We hypothesize that different domains share underlying neural pathways in the model's representation space. For example, factuality may share some representational structure with expert tone, while harmful content may overlap with certain emotional attributes. Understanding these shared structures allows for more effective control across multiple domains.

\subsection{Relationship to Model Alignment Mechanisms}
We hypothesize that various attributes and safety mechanisms share underlying neural pathways. If a model can be steered across multiple domains, similar techniques might be used to enhance or bypass alignment mechanisms that rely on similar representational structures. This has important implications for understanding the robustness of current alignment approaches.
\fi

\textbf{Train an MLP.} We use a multi-layer perceptron (MLP) for multi-label classification with input dimension $d_{\text{model}}$, two hidden layers (256 units, ReLU), and an output layer of size $K$ (softmax).

While we train the classifier using only the activations from the final position index at a given layer, when steering we apply it across all positions in the sequence. We train using the cross-entropy loss with the default settings of the PyTorch Adam optimizer, a batch size of 32, and train for 30 epochs, which is sufficient for the classifier to achieve high accuracy on our datasets.

\textbf{K-Steering: MLP-guided interventions.} 
We take the method proposed by \citet{casper_nonlinear} that employs non-linear binary probes to generate latent space adversarial attacks, and extend it to steer in multiple directions simultaneously. We show that we are able to influence model behavior toward or away from specific class characteristics. This technique operates by modifying activations at arbitrary layers through gradient-based optimization techniques.

We formalize the K-Steering problem as follows. Let $f_{\theta}$ be an autoregressive language model with parameters $\theta$. At inference time, $f_{\theta}$ produces activations ${\bf a}_i \in \mathbb{R}^{d_{seq} \times d_{model}}$ at a given layer, where $d_{seq}$ represents the sequence length and $d_{model}$ is the hidden dimension. Our classifier $g_{\phi}: \mathbb{R}^{d_{model}} \rightarrow \mathbb{R}^K$ with parameters $\phi$ maps these activations to $K$ categories (tones or debate styles). The K-Steering intervention can be formulated as:

\begin{equation}
\label{eq:k-steering-loss}
{\bf a}'_i = {\bf a}_i - \alpha \nabla_{{\bf a}_i} \mathcal{L}(g_{\phi}({\bf a}_i))
\end{equation}

where $\mathcal{L}$ is a loss function that maximizes the likelihood of target classes and minimizes the likelihood of classes to avoid, and $\alpha$ is a scaling factor. Given this formalization, we introduce two algorithms for K-Steering, as illustrated in Figure~\ref{fig:your_label}.

Algorithm~\ref{alg:gradient-iteration} operates by iteratively adjusting activations based on the gradient of a loss function that rewards high logit values for target classes and penalizes high values for classes to avoid. These logits are the raw, unnormalized outputs from the classifier network for each class. The updates are scaled by a decaying learning rate to ensure stability during optimization.

\begin{algorithm}
\caption{Iterative Gradient-based Steering}
\label{alg:gradient-iteration}
\begin{algorithmic}[1]
\STATE \textbf{Input:} Activation ${\bf a} \in \mathbb{R}^{d_{seq} \times d_{model}}$, target classes $T^+$, avoid classes $T^-$, initial learning rate $\alpha$, number of steps $K$, decay rate $\gamma$
\STATE ${\bf a}_0 = {\bf a}$
\FOR{$k = 0$ to $K-1$}
    \STATE $\alpha_k = \alpha \cdot \gamma^k$ \COMMENT{Apply learning rate decay}
    \STATE $\mathcal{L} = 0$
    \IF{$T^+$ is not empty}
        \STATE $\mathcal{L} = \mathcal{L} - \text{mean}(g_{\phi}({\bf a}_k)_{T^+})$ \COMMENT{Maximize logits for target classes}
    \ENDIF
    \IF{$T^-$ is not empty}
        \STATE $\mathcal{L} = \mathcal{L} + \text{mean}(g_{\phi}({\bf a}_k)_{T^-})$ \COMMENT{Minimize logits for avoid classes}
    \ENDIF
    \STATE Compute gradient $\nabla_{{\bf a}_k} \mathcal{L}$
    \STATE ${\bf a}_{k+1} = {\bf a}_k - \alpha_k \nabla_{{\bf a}_k} \mathcal{L}$
\ENDFOR
\STATE \textbf{Return:} ${\bf a}_K$
\end{algorithmic}
\end{algorithm}

It is important to note that when the activation ${\bf a}$ has dimensions $d_{seq} \times d_{model}$, the algorithm treats all sequence positions collectively. The $\text{mean}$ operation in lines 7 and 10 averages across both sequence positions and target/avoid classes, producing a single scalar loss. This design choice allows the steering to consider the entire sequence context when making updates, which is particularly appropriate for attributes like tone that are emergent properties of the entire sequence rather than individual tokens. The resulting gradient updates are applied to all positions. The hyperparameters for Algorithm~\ref{alg:gradient-iteration} are detailed in Appendix \ref{sec:hyperparameters-algorithm1}.

While Algorithm~\ref{alg:gradient-iteration} provides a general approach for simultaneously steering toward target classes and away from avoid classes, we also present Algorithm~\ref{alg:projection-removal}, which is specialized for scenarios where the primary goal is to remove undesired attributes. This method is conceptually related to directional ablation techniques established in prior work \citep{arditi2024refusal}, but leverages the gradient information from our non-linear classifier to dynamically identify the relevant direction for each specific context.

As shown in Algorithm~\ref{alg:projection-removal}, the projection removal process begins by computing the loss function as the mean of the logits for classes we want to avoid. The gradient of this loss with respect to the activation naturally points in the direction that would most strongly increase the classifier's prediction for the avoid tones. We then calculate the projection of the activation onto this gradient direction and remove it with an amplification factor.

\begin{algorithm}
\caption{Projection Removal}
\label{alg:projection-removal}
\begin{algorithmic}[1]
\STATE \textbf{Input:} Activation ${\bf a}$, avoid classes $T^-$
\STATE $\mathcal{L} = \text{mean}(g_{\phi}({\bf a})_{T^-})$ \COMMENT{Loss uses raw logits for avoid classes}
\STATE Compute gradient $\nabla_{{\bf a}} \mathcal{L}$
\STATE Compute norm $\|{\nabla_{{\bf a}} \mathcal{L}}\|^2$
\STATE Compute dot product $d = {\bf a} \cdot \nabla_{{\bf a}} \mathcal{L}$
\STATE Compute projection ${\bf p} = (d / \|{\nabla_{{\bf a}} \mathcal{L}}\|^2) \cdot \nabla_{{\bf a}} \mathcal{L}$
\STATE ${\bf a}' = {\bf a} - 2 \cdot {\bf p}$ 
\STATE \textbf{Return:} ${\bf a}'$
\end{algorithmic}
\end{algorithm}

The key operation in our approach (line 7 of Algorithm~\ref{alg:projection-removal}) is a Householder reflection \citep{householder}, of the activation vector across the hyperplane perpendicular to the gradient direction. This operation can be expressed as ${\bf a}' = {\bf a} - 2\frac{({\bf a} \cdot {\bf v})}{({\bf v} \cdot {\bf v})}{\bf v}$, where ${\bf v}$ is the normal vector to the hyperplane (in our case, the gradient). Unlike simple projection removal, this transformation reverses the component along the gradient direction, effectively pushing the activation away from the undesired attribute boundaries in the non-linear activation space.  This algorithm is computationally more efficient as it requires only a single gradient computation and no iterative updates. Our empirical results in Section \ref{sec:projection-removal-results} demonstrate that it outperforms directional ablation in the debates task.

\section{Datasets and models}
\label{sec:datasets}
\textbf{Model Selection.} For our experiments, we use the Llama-3.2-3B-Instruct \citep{llama3}, Mistral-7B-Instruct-v0.3 \citep{mistral7b} and OLMo-2-1124-7B-Instruct \citep{olmo2} models. These models provide a good balance between size and performance, allowing for efficient experimentation while still producing high-quality outputs. %We use the TransformerLens library \cite{nanda2022transformerlens} to access the model's internal activations and implement our steering interventions.

For our comparisons, we required datasets where we can (1) steer \textit{between} one attribute to another, (2) steer \text{away} from an attribute, and (3) steer towards composable combinations of attributes that compose together. The datasets used by prior work such as \citet{song2025effectively} and \citet{nguyen2025multi} are effectively combinations of binary datasets, and not ``truly" k attributes that are composable in a single response. To fill this gap, we use GPT-4o-Mini to compile two k attribute datasets to demonstrate the versatility of K-Steering. In the first case, we create 
\textsc{ToneBank}, a dataset of questions that can be responded to in six conversational tones, and \textsc{DebateMix} which can be responded to in ten rhetorical styles. For a list of the labels for \textsc{ToneBank}, see Appendix \ref{sec:tone-labels}. For a list of the labels for \textsc{DebateMix}, see Appendix \ref{sec:debate-labels}.

\section{Single Layer, One Step K-Steering}
\label{sec:single-step-single-layer}

\begin{figure*}[!t]
\centering
\includegraphics[width=0.85\textwidth]{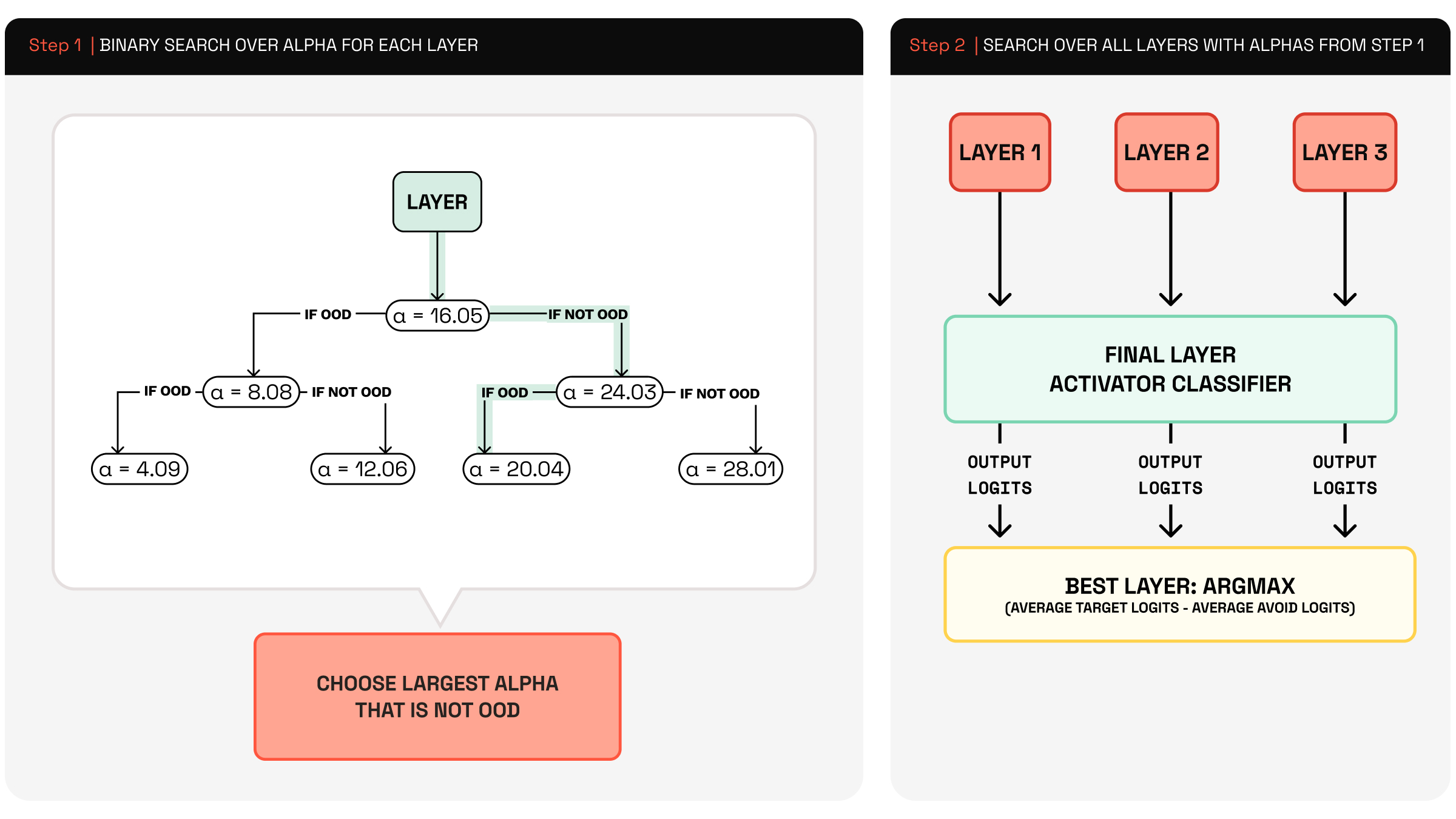}
\caption{Illustration of our evaluation setup for comparing CAA, DCT and K-Steering. In Step 1, we perform a binary search over $\alpha$ on the layers we want to steer. If we detect that an $\alpha$ causes the steered model to behave out-of-distribution, we decrease the alpha according to the binary search, otherwise we increase it. In Step 2, we use the alphas found through our binary search to select the best layer to steer on according to a classifier on the final layer activations. This is done by finding which layer when steered gives the highest mean probability across the target labels.}
\label{fig:evaluation}
\end{figure*}

\textbf{Evaluation.} We first experiment with a minimal implementation of K-Steering on a single layer. We intervene on the residual stream activations at one layer by taking a single gradient step in the direction of $K$ target labels according to the loss function described in Equation \ref{eq:k-steering-loss}, where $K$ is 1, 2 or 3. 

We compare K-Steering to Contrastive Activation Addition (CAA) \citep{caa} and Deep Causal Transcoding (DCT) \citep{dct} (for details on our baselines, see Appendix \ref{sec:baselines}). For CAA, we again intervene on the residual stream activations at a single point, computing a mean vector from 100 contrastive pairs. The positive element of a pair is the final token activations from a generation caused by a prompt instructing the model to respond according to a description of the target label. The negative element is the final token activations caused by prompting for any other label. For the labels and prompts used in each of our datasets, see Appendix \ref{sec:dataset_examples}. 

To create multi-label CAA vectors, we take the mean of single-label CAA vectors. If a label is to be avoided, we subtract the vector for that label before averaging. The DCT baseline learns 256 steering vectors over the last 5 layers of the model.

All three methods use a novel technique to calibrate the constants by which the steering vectors are scaled, which we refer to as $\alpha$s. A unique $\alpha$ is calibrated for every label combination through a 12-iteration binary search over a large range. A typical range is $[0.1, 1024]$. Although 1024 is much larger than most optimal $\alpha$ values, our binary search is computationally cheap and quickly rules out excessively large candidates. This ensures robustness against the edge case where performant $\alpha$s are very large.

The scoring function for our search is GPT-4o-Mini, which outputs an integer score between 0 and 100, representing the coherence of a generation steered by a vector scaled by some $\alpha$ (see Appendix \ref{sec:coherence_judge} for details). If the judge assigns a score above a threshold we search the upper half of the current range for a higher $\alpha$; otherwise, we search the lower half. The motivating heuristic that $\alpha$ should be as large as possible without affecting the model's coherence, ensuring the target behaviors are maximally visible. After 12 iterations, we select the largest $\alpha$ deemed coherent.

To find an appropriate layer to steer using CAA or K-Steering, we train a classifier on the activations of the final layer of a model. This classifier is identical in architecture and training to the classifiers used to compute K-Steering vectors, however it is trained on a heldout set of labelled activations and is never used to compute K-Steering gradient updates. The best layer is decided by the layer that has the largest positive mean difference from the unsteered model across all label combinations over 100 inputs. We depict the full evaluation in Figure \ref{fig:evaluation}. The use of a classifier on final layer activations addresses previous criticisms of evaluating steering methods based solely on outputs \citep{reliable_evaluation}.

To use the DCT vectors in our evaluation, we iterate through the 256 vectors, mapping each target label to a DCT vector according to which has the strongest positive effect on the classifier probabilities over 100 samples. When steering in multiple directions with DCT vectors, we use the mean of the vector for the individual target labels.

\textbf{Results.} We find that K-Steering outperforms CAA and DCT across all models and values of $K$, tying with DCT in the $K=3$ case on Mistral-7B on \textsc{ToneBank} (Table \ref{tab:combined_k_summary}). All three evaluated methods perform better than the unsteered model, except CAA in the $K=3$ case for Llama-3.2-3B on \textsc{ToneBank}. We additionally evaluate the performance of K-Steering, CAA and DCT at single attribute steering on TruthfulQA questions in Appendix \ref{sec:ksteer_single_attribute}.

To validate that our final layer classifier is giving reasonable scores, we select the label combinations with the largest difference in activation classifier score between K-Steering and CAA for each model, $K$, and dataset, finding whether an LLM judge agrees with the classifier judgement. We sample 100 generations from an unsteered model, a model steered with CAA, and a model steered with K-Steering, and prompt the judge to grade from 0 to 10 the extent to which the steered generations exhibit the target labels more so than the unsteered generations. The average of these scores gives a judgement for which method caused generations that exhibit the target labels more strongly. For more details on our judge, see Appendix \ref{sec:steering_judge}.

As seen in Table \ref{tab:llm_judge_scores}, the judge agreed with the classifier in 10 of 15 cases, suggesting agreement with the activation classifier. Our motivation for selecting the label combinations with the largest differences in activation classifier score was to confirm that the activation classifier was giving meaningful scores in the most extreme cases. Deviations of $<10\%$ (common when $K=3$) in the final layer activation classifier probabilities can be less visible in generations, which may not be as sensitive to steering as the residual stream activations.

\begin{table}[htbp]
\centering
\footnotesize
\begin{tabular}{@{}llccccc@{}}
\toprule
\textbf{$K$} & \textbf{Method} 
& \multicolumn{3}{c}{\textbf{\makecell{ToneBank\\(Olmo Llama Mistral)}}} 
& \multicolumn{2}{c}{\textbf{\makecell{DebateMix\\(Olmo Llama)}}} \\
\midrule
\multirow{3}{*}{1} 
& CAA   & 0.11 & 0.12 & 0.24 & 0.43 & 0.02 \\
& DCT   & 0.18 & 0.13 & 0.24 & 0.32 & 0.29 \\
& \textbf{K-Steer} & \textbf{0.37} & \textbf{0.17} & \textbf{0.46} & \textbf{0.56} & \textbf{0.56} \\
\addlinespace[0.5ex]
\multirow{3}{*}{2} 
& CAA   & 0.13 & 0.12 & 0.12 & 0.24 & 0.07 \\
& DCT   & 0.15 & 0.13 & 0.05 & 0.16 & 0.20 \\
& \textbf{K-Steer} & \textbf{0.21} & \textbf{0.17} & \textbf{0.09} & \textbf{0.25} & \textbf{0.25} \\
\addlinespace[0.5ex]
\multirow{3}{*}{3} 
& CAA   & 0.04 & 0.00 & 0.03 & 0.04 & 0.08 \\
& DCT   & 0.01 & 0.04 & \textbf{0.07} & 0.01 & 0.07 \\
& \textbf{K-Steer} & \textbf{0.09} & \textbf{0.07} & \textbf{0.07} & \textbf{0.09} & \textbf{0.09} \\
\bottomrule
\end{tabular}
\caption{Average change in final‐layer activation classifier probabilities induced by different steering methods, relative to an unsteered baseline, computed over all label combinations of size $K=1,2,3$. Results are shown for both the \textsc{ToneBank} and \textsc{DebateMix} datasets across different models.}
\label{tab:combined_k_summary}
\end{table}

\textbf{General Performance Benchmarking.} 
We carry out a benchmarking on a subset of MMLU under 3 shots for Olmo-2-7B and Llama-3.2-3B , curated by picking the first $3$ elements of each of the $57$ tasks. We find K-steering to outperform CAA across 1, 2 and 3 tone combinations, frequently matching the base model accuracy, suggesting that for at least some general benchmarks and parameters, K-Steering preserves language model performance competitively to CAA. See Table \ref{tab:benchmarking} for full results.

\begin{table}[htbp]
\centering
\footnotesize
\begin{tabular}{lcccc}
\toprule
\textbf{Model} & \textbf{K} & \textbf{K-Steering} & \textbf{CAA} & \textbf{Agree?} \\
\midrule
\multicolumn{5}{l}{\textbf{ToneBank}} \\
\multirow{3}{*}{Olmo-2-7B} & 1 & \textbf{0.22} & 0.16 & Y \\
                      & 2 & \textbf{0.26} & 0.14 & Y \\
                      & 3 & \textbf{0.28} & 0.26 & Y \\
\multirow{3}{*}{Llama-3.2-3B} & 1 & \textbf{0.28} & 0.07 & Y \\
                      & 2 & \textbf{0.24} & 0.23 & Y \\
                      & 3 & 0.29 & \textbf{0.33} & N \\
\multirow{3}{*}{Mistral-7B} & 1 & \textbf{0.37} & 0.36 & Y \\
                        & 2 & 0.25 & \textbf{0.27} & N \\
                        & 3 & 0.27 & \textbf{0.30} & N \\
\midrule
\multicolumn{5}{l}{\textbf{DebateMix}} \\
\multirow{3}{*}{Olmo-2-7B} & 1 & \textbf{0.51} & 0.36 & Y \\
                      & 2 & \textbf{0.28} & 0.25 & Y \\
                      & 3 & \textbf{0.30} & 0.11 & Y \\
\multirow{3}{*}{Llama-3.2-3B} & 1 & 0.37 & \textbf{0.38} & N \\
                      & 2 & \textbf{0.33} & 0.32 & Y \\
                      & 3 & 0.26 & \textbf{0.34} & N \\
\bottomrule
\end{tabular}
\caption{LLM judge scores (normalized to 0-1) for \textbf{CAA} and \textbf{K-Steering} across models, label set sizes ($K$), and datasets. We choose the label combinations with the highest divergence in activation classifier score between K-Steering and CAA. The \textbf{Agree?} column indicates whether final-layer activation classifier predicted the same winner.}
\label{tab:llm_judge_scores}
\end{table}

\begin{table}[htbp]
\centering
\footnotesize
\begin{tabular}{lcccc}
\toprule
\textbf{Model} & \textbf{K} & \textbf{K-Steering} & \textbf{CAA} & \textbf{Unsteered} \\
\midrule
\multirow{3}{*}{Olmo-2-7B} & 1 & \textbf{0.579} & \textbf{0.579} & \textbf{0.579} \\
                      & 2 & \textbf{0.579} & 0.575 & \textbf{0.579} \\
                      & 3 & \textbf{0.579} & 0.561 & \textbf{0.579} \\
\multirow{3}{*}{Llama-3.2-3B} & 1 & \textbf{0.573} & 0.540 & 0.573 \\
                      & 2 & \textbf{0.573} & 0.528 & \textbf{0.573} \\
                      & 3 & \textbf{0.573} & 0.555 & \textbf{0.573} \\
\bottomrule
\end{tabular}
\caption{Comparison of MMLU scores when models are steered with K-Steering, CAA, and are unsteered toward all combinations of attributes up to $K=3$ from \textsc{TONEBANK}. We give the mean MMLU score over all attribute combinations for a given value of $K$.}
\label{tab:benchmarking}
\end{table}

\begin{table}[htbp]
\centering
\begin{tabular}{lcc}
\toprule
\textbf{Combination} & \textbf{K-Steering} & \textbf{CAA} \\
\midrule
Burden/Reductio     & 4.2 / \textbf{0.80} & 0.14 / \textbf{0.80} \\
Burden/Strawman     & 9.4 / 0.72          & 0.18 / \textbf{0.80} \\
Empirical/Reductio  & 3.4 / \textbf{0.80} & 0.14 / \textbf{0.80} \\
Precedent/Burden    & 3.8 / \textbf{0.81} & 0.14 / 0.80 \\
Reductio/Analogy    & 10.2 / \textbf{0.80} & 0.22 / 0.68 \\
Reductio/Concession & 6.2 / 0.59          & 0.18 / \textbf{0.79} \\
\bottomrule
\end{tabular}
\caption{Best $\alpha$ / steering score for K-Steer and CAA across six style combinations. 6 randomly selected target/avoid style combination. We see that K-Steering wins across 50\% of the chosen combinations. The classifier is trained on layer 14 (middle layer)}
\label{tab:ksteer_caa_across_alphas}
\end{table}

\section{In-depth Analysis of K-Steering}
\label{sec:in-depth-analysis}
Given that single-step, single-layer K-Steering already outperforms baselines, we now examine whether additional gains can be achieved by modifying key intervention parameters. In particular, we investigate the effects of steering across multiple layers at once and taking multiple gradient steps.

\textbf{Multi-layer K-Steering.} We demonstrate that K-Steering can be effectively applied across multiple layers simultaneously. Our approach involves training a classifier at a specific layer $x$ in the residual stream, then leveraging this classifier to intervene across all residual stream layers. This multi-layer intervention enables stronger steering for complex tasks, such as transforming a model's reasoning from one debate style to another.

The efficacy of multi-layer steering aligns with established understanding of language model architecture. Models exhibit the "hydra effect"—self-repair mechanisms that resist single-point interventions \citep{mcgrath2023hydra}. Additionally, since information persists throughout the residual stream, specific latent representations likely appear across multiple layers \citep{lawson2024residual}, making multi-layer intervention both theoretically sound and empirically effective.

Our experiments demonstrate that both K-Steering and CAA achieve high steering scores when applied across all residual stream layers, with steering scores close to 0.8 (see Table \ref{tab:ksteer_caa_across_alphas}). Surprisingly, there is no clear winner. The steering scores reported in this section as well as subsequent ones use a comprehensive evaluation methodology that incorporates steering strength, success rate, and language model coherence, assessed by an LLM judge according to the rubric detailed in Appendix \ref{sec:steering_score}. To vary the steering strength, we explore a broad search over $\alpha \in [0.1, 30.0]$\footnote{We used fine-grained increments from 0.1 to 1.0, moderate steps from 1.0 to 10.0, and coarser steps up to 30.0 to efficiently span small and large intervention magnitudes.}, covering both subtle and large-scale intervention strengths.

We further demonstrate that classifier training can occur at various layers throughout the network, though with some variation in steering performance. As shown in Table \ref{tab:steering_scores_across_layers}, steering scores range from 0.522 to 0.860 across different layers, with layer 14 achieving the highest performance. Most notably, we observe that the optimal steering strength parameter $\alpha$ exhibits strong layer-specific dependencies—where earlier layers (e.g., layer 4) require substantially smaller $\alpha$ values (0.7) compared to deeper layers (e.g., layer 27) which perform best with much larger values (25.0).

\begin{figure}[htbp]
    \centering
    \includegraphics[width=0.49\textwidth]{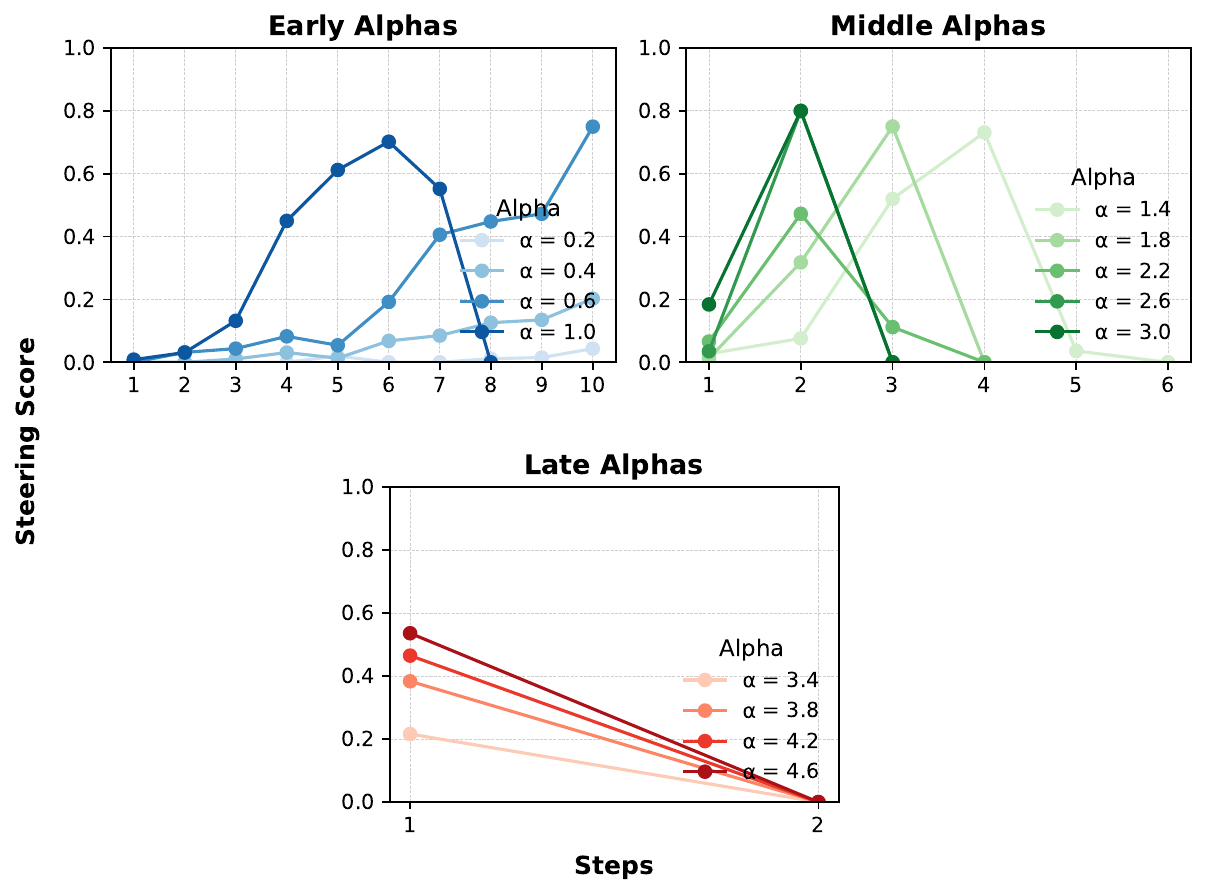}
    \caption{
        Steering scores across steps for 3 groups of $\alpha$ values when steering towards the "Empirical Grounding" style and away from the "Straw Man Reframing" style. We perform the experiment on layer 14 in Llama-3.2-3B.
        We divide $\alpha$ into 3 ranges: early ($\alpha \leq 1.0$), middle ($1.0 < \alpha \leq 3.0$), and late ($\alpha > 3.0$). 
        Each subplot shows how multi-step steering influences intervention effectiveness, with darker shades representing larger $\alpha$ values within each range.
        Highest performance is observed at $\alpha = 3.0$ and $\alpha = 2.6$ using two steps.
    }
    \label{fig:steering-score-alpha-groups}
\end{figure}

\textbf{Multi-step K-Steering.}
We next examine whether K-Steering performance can be improved by increasing the number of intervention steps. In this analysis, we vary both $\alpha$ and the number of steps. Single-step steering corresponds to a linear intervention along the tangent direction, while multi-step steering allows exploration of the non-linear loss landscape. As shown in Figure~\ref{fig:steering-score-alpha-groups}, multi-step steering generally improves effectiveness, especially for smaller $\alpha$ values, which benefit from a smoother, incremental gain across steps. In contrast, mid-range $\alpha$ values yield sharp performance peaks within a few steps, while large $\alpha$ values rapidly degrade model coherence, resulting in scores collapsing to zero.

In addition, we show that allowing multiple gradient steps enables us to search over a narrower $\alpha$ range\footnote{The narrower $\alpha$ range used for multi-step steering is $[0.2, 4.6]$ in uniform increments of 0.4.} while achieving comparable performance (see Table~\ref{tab:steering-scores-alphas-steps}) to single-step steering over a broader range (see Table~\ref{tab:ksteer_caa_across_alphas}). However, this performance gain comes at the cost of a linear increase in compute, proportional to the number of steps per input sample. See Appendix~\ref{sec:cost_comparison} for a detailed breakdown of the computational cost.

\begin{table}[htbp]
\centering
\resizebox{\columnwidth}{!}{%
\begin{tabular}{lccc}
\toprule
\textbf{Debate Style} & \textbf{Best $\alpha$} & \textbf{Steps} & \textbf{Score} \\
\midrule
Burden/Reductio     & 1.0 & 7 & 0.80 \\
Burden/Strawman     & 1.0 & 8 & 0.79 \\
Empirical/Reductio  & 0.6 & 8 & 0.80 \\
Precedent/Burden    & 0.6 & 6 & 0.77 \\
Reductio/Analogy    & 1.8 & 3 & 0.63 \\
Reductio/Concession & 0.6 & 7 & 0.57 \\
\bottomrule
\end{tabular}
}
\caption{Optimal $\alpha$, number of steps, and steering scores for multi-step K-Steering across debate style pairs. We perform the experiments on layer 14 in Llama-3.2-3B. }
\label{tab:steering-scores-alphas-steps}
\end{table}

For our multi-step steering experiments, we evaluate a fixed range of 1 to 10 steps. We observe that performance for small $\alpha$ values continues to improve across this range, and we believe that increasing the number of steps beyond 10 could yield further gains (see Figure~\ref{fig:steering-score-alpha-groups}). However, due to the increasing computational overhead, we leave this exploration to future work. We show some sampled K-steered completions in Appendix \ref{sec:k-steered-completions} with multi-step steered completions in Section \ref{sec:multi-step-steering-completions}.

\section{Projection removal}
\label{sec:projection-removal-results}
In this section, we evaluate the effectiveness of the projection removal technique described in Algorithm~2 on both \textsc{DebateMix} and \textsc{ToneBank} tasks. This method aims to steer the model away from a specific debate style or conversational tone by ablating gradient directions associated with the corresponding class logits. We compare it against the ``difference-in-means'' directional vector ablation baseline introduced by \citet{arditi2024refusal}, referred to as CAA.

As shown in Table~\ref{tab:projection-removal-results-debates-tones}, projection removal K-Steering outperforms CAA in 6 of 10 debate styles, with notable gains for burden, empirical, and refutation. While CAA performs better on most conversational tones, the differences are typically under 0.04 with the exception of the expert tone, where CAA has a larger lead. These results suggest that projection removal K-Steering is especially effective on the debates task and remains competitive for the tones task.

\section{Conclusion}
\label{sec:conclusion}
We introduced K-Steering, a unified method for multi-attribute control in language models using gradients from a non-linear classifier trained on hidden activations. Unlike prior approaches, K-Steering enables simultaneous control across multiple behaviors without relying on linear vector composition or per-attribute tuning.

Our main evaluation focuses on a single-step, single-layer variant, showing that K-Steering outperforms strong baselines on two new benchmarks: \textsc{ToneBank} and \textsc{DebateMix}. We also explore multi-step and multi-layer. While multi-step steering can enhance effectiveness, it introduces additional computational overhead.

Although our results highlight the flexibility and strength of classifier-guided steering, this work also raises challenges related to scalability, dataset construction, and baseline coverage, covered in the Limitations section. We hope this work provides a foundation for more robust compositional behavior control in LLMs.

\section*{Limitations}

Despite our best efforts, several limitations remain in the present work.

\noindent \textbf{Datasets.} Our experiments are conducted on two custom-constructed datasets designed specifically to test multi-dimensional steering. In naturally occurring datasets, behavior labels infrequently compose in a meaningful or consistent way, resulting in a scarcity of composable examples. As such, the generalizability of our method to other datasets remains an open question.

\noindent \textbf{Number of Steering Vectors.} The number of possible steering directions grows exponentially with the number of target behaviors (i.e., $\sum_{k=1}^n \binom{n}{k}$). To keep the experimental scope tractable, we restrict our evaluation to combinations involving at most three behaviors per dataset.

\noindent \textbf{More comprehensive general benchmarking performance.} Ideally, we would have carried out a more comprehensive evaluation of model performance across more models, hyperparameter and attribute combinations, given greater time and compute resources.

\noindent \textbf{Multi-step K-Steering.} Our current implementation applies multi-step steering sequentially for each input prompt. This results in a combinatorial explosion when searching over grid values of $\alpha$ and step counts, making the approach computationally expensive—several orders of magnitude more costly than baseline methods. Consequently, we were only able to evaluate multi-step K-Steering on a limited number of combinations.

\noindent \textbf{Baselines.} While we carefully tuned hyperparameters (e.g., steering layers and magnitudes) for both our method and all included baselines to ensure a fair comparison, we were unable to directly evaluate a broader range of baseline methods such as ITI\citep{inference_li} and RepE\cite{zou2023representation} due to computational constraints. We included CAA as this was by far the \emph{best} performing of the mainstream steering methods evaluated in \citet{unified}, shown to outperform ITI and RepE by large margins across a wide range of datasets.

We plan to address many of these limitations and extensions in our future work, see Appendix \ref{sec:future}.

\section*{Ethics Statement}
While K-Steering has potential beneficial applications for enhancing model controllability, it also raises ethical concerns. The ability to manipulate model outputs could be misused to bypass safety measures or to make models generate harmful content. We emphasize the importance of responsible use of these techniques and suggest the development of countermeasures to protect against potential misuse.

\section*{Acknowledgments}
We thank Michael Lan and Alice Riggs for their detailed feedback and suggestions, which significantly improved the clarity and presentation of this paper. We thank Antia Garcia Casal for designing Figures \ref{fig:your_label} and Figures \ref{fig:evaluation}. We also thank Shriyash Upadhyay, Jeff Phillips, Abir Harrasse, Nirmalendu Prakash, Chaithanya Bandi and Philip Quirke for their helpful discussions, perspective and feedback.

\bibliography{acl_latex}

\appendix
\section*{Appendix}

\label{sec:hyperparameters-algorithm1}

\section{Hyperparameters Algorithm~\ref{alg:gradient-iteration} }
The hyperparameters for Algorithm~\ref{alg:gradient-iteration} are as follows:
\begin{itemize}
    \item $\alpha$ (initial learning rate): Controls the magnitude of each gradient step. Higher values lead to larger updates but may overshoot optimal solutions, while smaller values provide more precise control while requiring more iterations. 
    
    \item $K$ (number of steps): The number of gradient descent iterations to perform. More steps allow for finer adjustments toward target tones but increase computational cost.
    
    \item $\gamma$ (decay rate): Rate at which the learning rate decreases with each iteration, calculated as $\alpha_k = \alpha \cdot \gamma^k$. Values closer to 1 maintain consistent step sizes, while smaller values rapidly reduce step sizes in later iterations for more stable convergence. 
\end{itemize}

\begin{table}[H]
\centering
\begin{tabular}{@{}llccc@{}}
\toprule
\textbf{Dataset} & \textbf{Avoid} & \textbf{CAA} & \textbf{K-Steer} & \textbf{DCT} \\
\midrule
\multicolumn{5}{l}{\textit{Debate Styles}} \\
& analogy       & 0.451 & \textbf{0.551} & 0.179 \\
& burden        & 0.413 & \textbf{0.684} & 0.364 \\
& circular      & 0.468 & 0.504 & \textbf{0.631} \\
& concession    & \textbf{0.486} & 0.338 & 0.322 \\
& empirical     & 0.570 & \textbf{0.690} & 0.471 \\
& moral         & \textbf{0.620} & 0.570 & 0.374 \\
& precedent     & \textbf{0.630} & 0.513 & 0.462 \\
& reductio      & 0.432 & \textbf{0.627} & 0.392 \\
& refutation    & 0.567 & \textbf{0.700} & 0.096 \\
& strawman      & \textbf{0.254} & 0.100 & 0.144 \\
\midrule
\multicolumn{5}{l}{\textit{Conversational Tones}} \\
& casual        & \textbf{0.800} & 0.770 & 0.175 \\
& cautious      & \textbf{0.790} & 0.750 & 0.493 \\
& concise       & \textbf{0.690} & 0.670 & 0.620 \\
& empathetic    & \textbf{0.790} & \textbf{0.790} & 0.612 \\
& expert        & \textbf{0.703} & 0.428 & 0.563 \\
& helpful       & \textbf{0.740} & 0.720 & 0.458 \\
\bottomrule
\end{tabular}
\caption{Comparison of steering scores using Projection Removal (K-Steering), Directional Ablation (CAA), and the new DCT method across debate styles and conversational tones. Bolded values indicate the best method per row. DCT scores computed as $(\text{average\_strength}/5)\times\text{success\_rate}$. We perform the experiment on layer 14 in Llama-3.2-3B.}
\label{tab:projection-removal-results-debates-tones}
\end{table}

\section{Steering scores distribution across layers}
\begin{table}[H]
\centering
\begin{tabular}{ccc}
\toprule
Layer & Best $\alpha$ & Score \\
\midrule
4 & 0.7 & 0.740 \\
7 & 1.0 & 0.522 \\
14 & 4.5 & \textbf{0.860} \\
27 & 25.0 & 0.648 \\
\bottomrule
\end{tabular}
\caption{Best steering score across layers (Precedent vs Reductio). We report the steering scores across four uniformly distributed layers of LLama-3.2-3B for a randomly selected steering combination pair (steering towards Precedent and away from Reductio).}
\label{tab:steering_scores_across_layers}
\end{table}

\section{Computational Cost Comparison}
\label{sec:cost_comparison}

We compare the computational cost of our iterative gradient-based steering method with the contrastive activation addition (CAA) baseline. Let:

\begin{itemize}
    \item $d_{\text{seq}}$: sequence length
    \item $d_{\text{model}}$: activation dimensionality
    \item $H$: hidden layer size of the MLP classifier
    \item $C$: number of tone classes
    \item $N$: number of steering steps
\end{itemize}

\subsection{Gradient-Based Steering Cost}

For each input sample, our method performs $N$ iterations of gradient-based steering. Each iteration consists of:
\begin{enumerate}
    \item A forward pass through a 3-layer MLP:
    \[
    \mathcal{O}\left(d_{\text{seq}} \cdot (d_{\text{model}} \cdot H + H^2 + H \cdot C)\right)
    \]
    \item A backward pass to compute gradients w.r.t. the input activation (same cost as forward).
    \item An activation update:
    \[
    \mathcal{O}(d_{\text{seq}} \cdot d_{\text{model}})
    \]
\end{enumerate}

These steps are repeated independently for each of the $N$ iterations, with no reuse of computation between steps. This is because each iteration performs a new forward and backward pass based on the updated activation vector, followed by a gradient descent step. As a result, the total cost scales linearly with $N$.

The total cost over $N$ steps is therefore:
\[
\boxed{
\mathcal{O}\left(N \cdot d_{\text{seq}} \cdot (d_{\text{model}} \cdot H + H^2 + H \cdot C)\right)
}
\]

\subsection{Contrastive Activation Addition (CAA) Cost}

CAA performs a single vector addition per activation position:
\[
\boxed{
\mathcal{O}(d_{\text{seq}} \cdot d_{\text{model}})
}
\]

\subsection{Comparison Summary}

\begin{table}[H]
\begin{tabular}{l p{0.30\textwidth}}
\toprule
\textbf{Method} & \textbf{Inference Cost per Sample} \\
\midrule
K-Steer & $\mathcal{O}\left(N \cdot d_{\text{seq}} \cdot \left(d_{\text{model}} \cdot H + H^2 \right.\right.$ \\
        & $\hspace{1.2em}\left.\left. + H \cdot C \right)\right)$ \\
CAA & $\mathcal{O}(d_{\text{seq}} \cdot d_{\text{model}})$ \\
\bottomrule
\end{tabular}
\caption{Asymptotic cost of steering methods at inference time.}
\end{table}

Our method provides fine-grained, class-aware control of steering directions at the cost of several orders of magnitude more computation per sample compared to CAA. The linear dependence on $N$ arises from the fact that each step independently computes a new gradient and updates the activation, with no amortization of costs across steps.

\section{Single-Attribute Steering Evaluation With Conventional Attributes}
\label{sec:ksteer_single_attribute}
In addition to our comparison of K-Steering with CAA and DCT using our multi-attribute datasets, we compare K-Steering, CAA and DCT when steering toward attributes that have historically been used to evaluate steering methods, such as refusal and bias. We perform our evaluation from Section \ref{sec:single-step-single-layer} on TruthfulQA questions.

We compute steering vectors for bias, refusal, toxicity and unhelpfulness for each of the methods and select the layers and alphas to steer with using the procedure in Section \ref{sec:single-step-single-layer}. The final layer evaluation classifier results are in Table \ref{tab:ksteer_single_attribute}.

\begin{table*}[htbp]
\centering
\begin{tabular}{lccc}
\toprule
\textbf{Attribute} & \textbf{K-Steering (\%)} & \textbf{CAA (\%)} & \textbf{DCT (\%)} \\
\midrule
Bias & \textbf{0.52} & 0.36 & 0.04 \\
Refusal & \textbf{0.43} & 0.37 & 0.11 \\
Toxicity & \textbf{0.52} & 0.41 & 0.35 \\
Unhelpfulness & \textbf{0.57} & 0.49 & 0.31 \\
\bottomrule
\end{tabular}
\caption{Final layer activation classifier scores caused by steering with different methods on TruthfulQA questions.}
\label{tab:ksteer_single_attribute}
\end{table*}

\section{Dataset creation}
\label{sec:dataset_creation}

\textbf{DEBATEMIX}. 
We use the following prompt to instruct GPT-4 (gpt4-o1-mini) to create a set of 978 prompts that can be responded to in one of the 10 debate styles. We use a temperature of 0.7 with a max of 800 tokens and ask the model to create between 30 and 60 examples per general category.

\begin{tcolorbox}[breakable, colback=gray!10, colframe=gray!50, arc=4pt, boxrule=0.4pt]
\small
\ttfamily
Generate \{prompts\_per\_category\} diverse, high-quality debate questions that fall under the category: ``\{category\}".\newline

IMPORTANT: Create questions that could be meaningfully approached using multiple different debate styles, such as:\newline
- Reductio ad Absurdum: Extending an opponent's argument to absurdity to show its flaws.\newline
- Appeal to Precedent: Justifying a stance by referring to a historical example or past decision.\newline
- Straw Man Reframing: Recasting an opponent's argument into a weaker version, then refuting it.\newline
- Burden of Proof Shift: Placing the obligation to prove or disprove a claim on the opponent.\newline
- Analogy Construction: Drawing parallels to a more familiar concept to clarify or persuade.\newline
- Concession and Pivot: Conceding a minor point to appear fair, then redirecting to a stronger argument.\newline
- Empirical Grounding: Citing specific data, studies, or real-world outcomes.\newline
- Moral Framing: Placing the argument within an ethical or moral framework.\newline
- Refutation by Distinction: Showing that an opponent's example or principle doesn't apply to the case at hand.\newline
- Circular Anticipation: Preempting and dismantling an opponent's likely response before they make it.\newline

Each generated prompt should:\newline
1. Be a clear and well-formed debatable question or statement\newline
2. Be style-neutral (able to be approached well using any of the debate styles)\newline
3. Have sufficient complexity to allow for nuanced arguments\newline
4. Avoid numbering or special formatting\newline
5. Be suitable for formal debate settings\newline

Focus on creating questions where the SAME question can be approached in meaningfully different ways depending on which debate style is used to argue the position.\newline

These should be questions where reasonable people might disagree, and where multiple debate techniques could be effectively employed.
\end{tcolorbox}

We show the distribution of prompts by category in Table \ref{tab:count_debates_by_category}.

\begin{table}[htbp]
\centering
\begin{tabular}{l r}
\toprule
\textbf{Category} & \textbf{Count} \\
\midrule
civil\_liberties & 34 \\
human\_rights & 37 \\
science & 40 \\
constitutional\_law & 42 \\
ethics & 45 \\
religious\_freedom & 47 \\
economics & 49 \\
education\_reform & 50 \\
healthcare\_systems & 51 \\
media\_influence & 55 \\
business\_regulation & 56 \\
technology\_impact & 57 \\
environmental\_policy & 57 \\
historical\_interpretation & 58 \\
philosophy & 58 \\
social\_issues & 59 \\
politics & 60 \\
cultural\_values & 60 \\
international\_relations & 63 \\
\bottomrule
\end{tabular}
\caption{DEBATEMIX rows by category}
\label{tab:count_debates_by_category}
\end{table}

\textbf{TONEBANK} is created by a similar process, and consists of 1184 examples distributed over 18 categories.

\begin{tcolorbox}[breakable, colback=gray!10, colframe=gray!50, arc=4pt, boxrule=0.5pt]
\small
\ttfamily
Generate \{prompts\_per\_category\} diverse, high-quality questions that fall under the category: "\{category\}".\newline

IMPORTANT: Create questions that could be meaningfully answered in multiple different tones, such as:\newline
- expert: technically precise with domain terminology\newline
- casual: conversational and friendly\newline
- cautious: highlighting limitations and uncertainties\newline
- empathetic: emotionally attuned and validating\newline
- concise: minimalist and efficient\newline

Each generated prompt should:\newline
1. Be a clear and well-formed question ending with a question mark\newline
2. Be tone-neutral (able to be answered well in any of the tones)\newline
3. Avoid numbering or special formatting\newline

Focus on creating questions where the SAME question can receive meaningfully different responses depending on which tone is used to answer.
\end{tcolorbox}

We include a count of example by category in Table \ref{tab:count_tones_by_category}.

\begin{table}[htbp]
\centering
\begin{tabular}{l r}
\toprule
\textbf{Category} & \textbf{Count} \\
\midrule
opinion\_requests & 55 \\
professional\_help & 57 \\
decision\_making & 59 \\
relationship\_advice & 60 \\
parenting\_questions & 63 \\
emotional\_support & 64 \\
mental\_health & 64 \\
career\_guidance & 64 \\
personal\_finance & 64 \\
personal\_advice & 65 \\
technical\_questions & 65 \\
creative\_requests & 66 \\
physical\_fitness & 67 \\
nutrition\_advice & 70 \\
life\_skills & 72 \\
how\_to\_guides & 74 \\
education\_inquiries & 76 \\
factual\_information & 79 \\
\bottomrule
\end{tabular}
\caption{TONEBANK rows by Category}
\label{tab:count_tones_by_category}
\end{table}

\section{Dataset Labels}
\label{sec:labels}
\subsection{ToneBank}
\label{sec:tone-labels}
\textsc{ToneBank}: We select six diverse tone categories, described for language model prompting as below:
\begin{enumerate}[itemsep=0pt, parsep=0pt, topsep=0pt, partopsep=0pt]
    \item \textbf{Expert:} formal, authoritative, using technical terminology
    \item \textbf{Empathetic:} warm, supportive, focusing on emotional understanding
    \item \textbf{Cautious:} hedging, acknowledging limitations, presenting multiple perspectives
    \item \textbf{Casual:} conversational, informal, using colloquial language
    \item \textbf{Concise:} brief, minimal, avoiding elaboration
\end{enumerate}

\subsection{DebateMix}
\label{sec:debate-labels}
\textsc{DebateMix}: We construct a dataset of debate questions that can be answered using the following ten styles:

\begin{enumerate}[itemsep=0pt, parsep=0pt, topsep=0pt, partopsep=0pt]
    \item \textbf{Reductio ad Absurdum:} Extend opponent's logic to absurd extremes to reveal flaws.
    \item \textbf{Appeal to Precedent:} Cite past rulings or history to justify present stance.
    \item \textbf{Straw Man Reframing:} Oversimplify opponent's view to refute an easier version.
    \item \textbf{Burden of Proof Shift:} Demand opponent disprove your claim to shift burden.
    \item \textbf{Analogy Construction:} Use relatable analogies to clarify and support your point.
    \item \textbf{Concession and Pivot:} Concede a minor point, then redirect to stronger arguments.
    \item \textbf{Empirical Grounding:} Rely on data, studies, and statistics to support your case.
    \item \textbf{Moral Framing:} Frame issue in terms of ethics and moral values.
    \item \textbf{Refutation by Distinction:} Highlight key differences that invalidate opponent's logic.
    \item \textbf{Circular Anticipation:} Preemptively address and rebut expected counterarguments.
\end{enumerate}

These are classical rhetoric and logical techniques, refer to \citet{debate_1, debate_2} for more details. We describe the creation of both datasets in Appendix \ref{sec:dataset_creation}.
We give the full prompts used to direct models to respond in these debate and tone styles in Appendix \ref{sec:dataset_examples},
alongwith selected prompts and completions.

\section{Dataset Examples}
\label{sec:dataset_examples}

We provide a snapshot of representative prompts from DEBATEMIX in Table~\ref{tab:prompt-snapshot-debtaes}, grouped by category and shown with a placeholder for the style-specific instruction appended before inputting them to the LLM.

\begin{table*}[t]
\centering
\footnotesize
\caption{Representative prompts from each category in the dataset, shown with a placeholder style instruction.}
\begin{tabular}{p{3cm} p{11cm}}
\toprule
\textbf{Category} & \textbf{Prompt} \\
\midrule
\textbf{business\_regulation} & ⟨RESPOND USING STYLE X⟩ Is it necessary for governments to regulate the use of facial recognition technology by businesses? \\
\textbf{civil\_liberties} & ⟨RESPOND USING STYLE X⟩ Should the government have the ability to detain individuals without trial in terrorism-related cases? \\
\textbf{constitutional\_law} & ⟨RESPOND USING STYLE X⟩ Is the procedure for amending the Constitution too rigid? \\
\textbf{cultural\_values} & ⟨RESPOND USING STYLE X⟩ Is the concept of cultural identity becoming outdated in a globalized world? \\
\textbf{economics} & ⟨RESPOND USING STYLE X⟩ Is free trade ultimately detrimental to national interests? \\
\textbf{education\_reform} & ⟨RESPOND USING STYLE X⟩ Should standardized tests be abolished? \\
\textbf{environmental\_policy} & ⟨RESPOND USING STYLE X⟩ Should corporations disclose their carbon footprints publicly? \\
\textbf{ethics} & ⟨RESPOND USING STYLE X⟩ Is it ethical to prioritize the majority over the rights of a minority? \\
\textbf{healthcare\_systems} & ⟨RESPOND USING STYLE X⟩ Can price transparency improve healthcare decision-making? \\
\textbf{historical\_interpretation} & ⟨RESPOND USING STYLE X⟩ Should we challenge established historical narratives? \\
\textbf{human\_rights} & ⟨RESPOND USING STYLE X⟩ Should leaders be held accountable for human rights violations? \\
\textbf{international\_relations} & ⟨RESPOND USING STYLE X⟩ Should countries prioritize their economic interests in climate talks? \\
\textbf{media\_influence} & ⟨RESPOND USING STYLE X⟩ Does media coverage of immigrants affect public opinion? \\
\textbf{philosophy} & ⟨RESPOND USING STYLE X⟩ Does justice require retribution, or can restoration suffice? \\
\textbf{politics} & ⟨RESPOND USING STYLE X⟩ Are political ideologies evolving in response to global challenges? \\
\textbf{religious\_freedom} & ⟨RESPOND USING STYLE X⟩ How do we balance religious freedom with social cohesion? \\
\textbf{science} & ⟨RESPOND USING STYLE X⟩ Can we balance innovation with privacy rights? \\
\textbf{social\_issues} & ⟨RESPOND USING STYLE X⟩ Should schools focus more on creativity and critical thinking? \\
\textbf{technology\_impact} & ⟨RESPOND USING STYLE X⟩ Should companies monitor employee productivity using technology? \\
\bottomrule
\end{tabular}
\label{tab:prompt-snapshot-debtaes}
\end{table*}

Tables~\ref{tab:prompt-completion-debate-1}--\ref{tab:prompt-completion-debate-12} present example prompts and corresponding completions across all debate styles. Each table illustrates how the model responds when steered using a specific rhetorical instruction. For each example, we show the full prompt—including the style-specific guidance—and the model's completion, highlighting the distinct rhetorical patterns elicited by each style.

\begin{table*}[ht]
\centering
\begin{footnotesize}
\caption{Prompt and completions for question 1 (styles 1–4).}
\label{tab:prompt-completion-debate-1}
\begin{tabular}{lll}
\toprule
\textbf{Style} & \textbf{Example \#} & \textbf{Prompt and Completion} \\
\midrule
\textbf{reductio} & 1 & \begin{minipage}[t]{0.65\textwidth}\footnotesize \textbf{Prompt:} RESPOND USING REDUCTIO AD ABSURDUM DEBATE STYLE. Extend the opposing viewpoint to its logical extreme to reveal its flaws. Demonstrate how the opposing position leads to absurd or contradictory conclusions when taken to its logical end. Structure your response to systematically expose the inconsistencies or impracticalities. Use conditional statements like 'If we follow this logic, then...' to illustrate the path to absurdity. Maintain a logical flow that shows how the initial premise inevitably leads to an unacceptable or ridiculous outcome.
Is the government's ability to impose restrictions on free speech justified in instances of misinformation that could lead to public harm? \\ \textbf{Completion:} To address this question, let's assume the opposing viewpoint that the government's ability to impose restrictions on free speech is justified in instances of misinformation that could lead to public harm. If we follow this logic, then:

1. **The government has the authority to dictate what information is true and what is not**: This implies \end{minipage} \\[0.8em] \midrule
\textbf{precedent} & 1 & \begin{minipage}[t]{0.65\textwidth}\footnotesize \textbf{Prompt:} RESPOND USING APPEAL TO PRECEDENT DEBATE STYLE. Ground your arguments in historical examples, established case law, or widely accepted previous decisions. Reference specific historical situations, court rulings, scientific discoveries, or cultural precedents that support your position. Draw explicit connections between these precedents and the current issue. Emphasize how respecting established precedents provides consistency and reliability. Point out the risks or inconsistencies of ignoring relevant precedents. Structure your response to show how past experiences inform present considerations.
Is the government's ability to impose restrictions on free speech justified in instances of misinformation that could lead to public harm? \\ \textbf{Completion:} To address this question, we must consider the historical context of government regulation of free speech, particularly in instances where misinformation poses a risk to public harm. The landmark case of Schenck v. United States (1919) provides a crucial precedent in this regard.

In Schenck, the Supreme Court established the " \end{minipage} \\[0.8em] \midrule
\textbf{strawman} & 1 & \begin{minipage}[t]{0.65\textwidth}\footnotesize \textbf{Prompt:} RESPOND USING STRAW MAN REFRAMING DEBATE STYLE. Recharacterize the opposing argument in simplified or exaggerated terms that are easier to refute. Present this reframed version as if it represents the original position. Systematically dismantle this reframed argument while implying you've addressed the original point. Use phrases like 'Essentially, what you're saying is...' or 'This position boils down to...' before presenting the simplified version. Follow with a thorough refutation of this reframed position. Conclude by suggesting that your refutation applies to the original argument as well.
Is the government's ability to impose restrictions on free speech justified in instances of misinformation that could lead to public harm? \\ \textbf{Completion:} Essentially, what you're saying is that the government should have the power to silence anyone who spreads misinformation, regardless of the context or intent, as long as it could potentially harm the public.

This position boils down to the idea that the government should be the sole arbiter of what is and isn't considered "h \end{minipage} \\[0.8em] \midrule
\textbf{burden} & 1 & \begin{minipage}[t]{0.65\textwidth}\footnotesize \textbf{Prompt:} RESPOND USING BURDEN OF PROOF SHIFT DEBATE STYLE. Redirect the responsibility for providing evidence to your opponent rather than proving your own claim. Challenge your opponent to disprove your assertion rather than supporting it yourself. Use phrases like 'There's no evidence that disproves...' or 'Can you definitively prove that isn't the case?' Position your claims as the default assumption that should be accepted until disproven. Question the sufficiency or quality of opposing evidence while demanding an impossibly high standard of proof. Emphasize that the lack of counter-evidence strengthens your position.
Is the government's ability to impose restrictions on free speech justified in instances of misinformation that could lead to public harm? \\ \textbf{Completion:} I'd like to shift the burden of proof to my opponent. Can they definitively prove that the government's ability to impose restrictions on free speech is never justified in instances of misinformation that could lead to public harm? I'm not aware of any instance where the government's restrictions on free speech have been deemed unjustified by \end{minipage} \\[0.8em] \midrule
\bottomrule
\end{tabular}
\end{footnotesize}
\end{table*}

\begin{table*}[ht]
\centering
\begin{footnotesize}
\caption{Prompt and completions for question 1 (styles 5–8).}
\label{tab:prompt-completion-debate-2}
\begin{tabular}{lll}
\toprule
\textbf{Style} & \textbf{Example \#} & \textbf{Prompt and Completion} \\
\midrule
\textbf{analogy} & 1 & \begin{minipage}[t]{0.65\textwidth}\footnotesize \textbf{Prompt:} RESPOND USING ANALOGY CONSTRUCTION DEBATE STYLE. Develop a vivid, relatable comparison between the complex issue at hand and something more familiar or intuitive. Build your argument around this carefully constructed parallel situation. Highlight specific points of similarity that support your position while addressing potential dissimilarities. Use phrases like 'This situation is similar to...' or 'To understand this concept, consider...' Ensure your analogy simplifies the complex issue without distorting its essential nature. Use the familiar scenario to guide your audience to your desired conclusion about the original issue.
Is the government's ability to impose restrictions on free speech justified in instances of misinformation that could lead to public harm? \\ \textbf{Completion:} The government's ability to impose restrictions on free speech in instances of misinformation that could lead to public harm is a complex issue that can be likened to a fire alarm system in a crowded theater. Just as a fire alarm is designed to alert people to a potential danger, the government's restrictions on free speech can be seen \end{minipage} \\[0.8em] \midrule
\textbf{concession} & 1 & \begin{minipage}[t]{0.65\textwidth}\footnotesize \textbf{Prompt:} RESPOND USING CONCESSION AND PIVOT DEBATE STYLE. Begin by acknowledging a minor point or critique from the opposing side to establish fairness and reasonableness. Use phrases like 'While it's true that...' or 'I can concede that...' followed by 'However,' 'Nevertheless,' or 'That said,' to redirect to your stronger arguments. Ensure the conceded point is peripheral rather than central to your main argument. After the concession, pivot decisively to your strongest points with increased emphasis. Frame your pivot as providing necessary context or a more complete perspective. Use the concession to demonstrate your objectivity before delivering your more powerful counterarguments.
Is the government's ability to impose restrictions on free speech justified in instances of misinformation that could lead to public harm? \\ \textbf{Completion:} I can concede that the government's primary concern in instances of misinformation is to protect the public from harm. While it's true that the government's role in regulating speech can be a delicate balance between freedom of expression and public safety, I acknowledge that there are situations where restrictions on free speech may be necessary to prevent harm. \end{minipage} \\[0.8em] \midrule
\textbf{empirical} & 1 & \begin{minipage}[t]{0.65\textwidth}\footnotesize \textbf{Prompt:} RESPOND USING EMPIRICAL GROUNDING DEBATE STYLE. Base your arguments primarily on verifiable data, research studies, statistics, and observable outcomes rather than theory or rhetoric. Cite specific figures, percentages, study results, or historical outcomes that support your position. Present evidence in a methodical manner, explaining how each piece of data relates to your argument. Address the reliability and relevance of your sources and methods. Compare empirical results across different contexts or time periods to strengthen your case. Anticipate and address potential methodological criticisms of the evidence you present.
Is the government's ability to impose restrictions on free speech justified in instances of misinformation that could lead to public harm? \\ \textbf{Completion:} To address this question, let's examine the empirical evidence on the impact of government restrictions on free speech versus the potential harm caused by misinformation.

**Government Restrictions on Free Speech:**

1. **Effectiveness in Reducing Misinformation:** A study published in the Journal of Communication (2019) found that government-mand \end{minipage} \\[0.8em] \midrule
\textbf{moral} & 1 & \begin{minipage}[t]{0.65\textwidth}\footnotesize \textbf{Prompt:} RESPOND USING MORAL FRAMING DEBATE STYLE. Position the issue within a framework of ethical principles, values, and moral imperatives rather than pragmatic concerns. Identify the core moral values at stake such as justice, liberty, equality, compassion, or responsibility. Use language that evokes ethical considerations, such as 'obligation,' 'right,' 'wrong,' 'just,' or 'fair.' Appeal to widely held moral intuitions or principles. Present opposing views as morally questionable or inconsistent with important shared values. Elevate the discussion from practical matters to questions of what ought to be done. Emphasize moral consequences over practical outcomes.
Is the government's ability to impose restrictions on free speech justified in instances of misinformation that could lead to public harm? \\ \textbf{Completion:} The government's ability to impose restrictions on free speech is a contentious issue that raises fundamental questions about the balance between individual liberties and the greater good. While the protection of free speech is a cornerstone of democratic societies, the government's duty to safeguard public health and prevent harm must also be considered.

In instances of misinformation that could \end{minipage} \\[0.8em] \midrule
\bottomrule
\end{tabular}
\end{footnotesize}
\end{table*}

\begin{table*}[ht]
\centering
\begin{footnotesize}
\caption{Prompt and completions for question 1 (styles 9–10).}
\label{tab:prompt-completion-debate-3}
\begin{tabular}{lll}
\toprule
\textbf{Style} & \textbf{Example \#} & \textbf{Prompt and Completion} \\
\midrule
\textbf{refutation} & 1 & \begin{minipage}[t]{0.65\textwidth}\footnotesize \textbf{Prompt:} RESPOND USING REFUTATION BY DISTINCTION DEBATE STYLE. Identify crucial differences that invalidate comparisons or principles your opponent has applied. Carefully delineate categories, contexts, or circumstances that demonstrate why a general rule or example doesn't apply in this specific case. Use phrases like 'While that may be true in some contexts...' or 'We must distinguish between...' Emphasize the precision of definitions and classifications. Highlight subtle but significant differences that undermine the opponent's logic. Show how these distinctions fundamentally change the assessment of the situation. Demonstrate how recognizing these distinctions leads to a different conclusion than your opponent reached.
Is the government's ability to impose restrictions on free speech justified in instances of misinformation that could lead to public harm? \\ \textbf{Completion:} While the government's ability to impose restrictions on free speech is a complex issue, I argue that it is not justified in instances of misinformation that could lead to public harm. My opponent may argue that the government has a responsibility to protect the public from harm, but I would counter that this responsibility is not a justification for suppressing \end{minipage} \\[0.8em] \midrule
\textbf{circular} & 1 & \begin{minipage}[t]{0.65\textwidth}\footnotesize \textbf{Prompt:} RESPOND USING CIRCULAR ANTICIPATION DEBATE STYLE. Preemptively identify and address the most likely counterarguments before your opponent can make them. Introduce opposing points with phrases like 'Some might argue...' or 'One could object that...' followed by your prepared refutation. Structure your response to cover all major potential objections. Demonstrate that you've thoroughly considered the issue from multiple angles. Frame potential counterarguments in ways that make them easier to dismantle. Create the impression that all reasonable objections have already been considered and overcome. Conclude by suggesting that any remaining objections would be similarly flawed.
Is the government's ability to impose restrictions on free speech justified in instances of misinformation that could lead to public harm? \\ \textbf{Completion:} The notion that the government's ability to impose restrictions on free speech is justified in instances of misinformation that could lead to public harm is a contentious issue. Some might argue that such restrictions infringe upon the fundamental right to free speech, and that the government should not have the power to dictate what information is disseminated to the \end{minipage} \\[0.8em] \midrule
\bottomrule
\end{tabular}
\end{footnotesize}
\end{table*}

\begin{table*}[ht]
\centering
\begin{footnotesize}
\caption{Prompt and completions for question 2 (styles 1–4).}
\label{tab:prompt-completion-debate-4}
\begin{tabular}{lll}
\toprule
\textbf{Style} & \textbf{Example \#} & \textbf{Prompt and Completion} \\
\midrule
\textbf{reductio} & 2 & \begin{minipage}[t]{0.65\textwidth}\footnotesize \textbf{Prompt:} RESPOND USING REDUCTIO AD ABSURDUM DEBATE STYLE. Extend the opposing viewpoint to its logical extreme to reveal its flaws. Demonstrate how the opposing position leads to absurd or contradictory conclusions when taken to its logical end. Structure your response to systematically expose the inconsistencies or impracticalities. Use conditional statements like 'If we follow this logic, then...' to illustrate the path to absurdity. Maintain a logical flow that shows how the initial premise inevitably leads to an unacceptable or ridiculous outcome.

Should the responsibilities of raising children be equally shared between parents? \\ \textbf{Completion:} To address the question of whether the responsibilities of raising children should be equally shared between parents, let's consider the opposing viewpoint and extend it to its logical extreme.

Assuming that the responsibilities of raising children should be equally shared between parents, we can infer that both parents should have an equal say in all major decisions regarding the \end{minipage} \\[0.8em] \midrule
\textbf{precedent} & 2 & \begin{minipage}[t]{0.65\textwidth}\footnotesize \textbf{Prompt:} RESPOND USING APPEAL TO PRECEDENT DEBATE STYLE. Ground your arguments in historical examples, established case law, or widely accepted previous decisions. Reference specific historical situations, court rulings, scientific discoveries, or cultural precedents that support your position. Draw explicit connections between these precedents and the current issue. Emphasize how respecting established precedents provides consistency and reliability. Point out the risks or inconsistencies of ignoring relevant precedents. Structure your response to show how past experiences inform present considerations.

Should the responsibilities of raising children be equally shared between parents? \\ \textbf{Completion:} The notion that the responsibilities of raising children should be equally shared between parents is a widely debated topic. However, a closer examination of historical precedents and established case law reveals that this approach is not only impractical but also inconsistent with the complexities of family dynamics.

One notable example is the landmark case of Troxel v \end{minipage} \\[0.8em] \midrule
\textbf{strawman} & 2 & \begin{minipage}[t]{0.65\textwidth}\footnotesize \textbf{Prompt:} RESPOND USING STRAW MAN REFRAMING DEBATE STYLE. Recharacterize the opposing argument in simplified or exaggerated terms that are easier to refute. Present this reframed version as if it represents the original position. Systematically dismantle this reframed argument while implying you've addressed the original point. Use phrases like 'Essentially, what you're saying is...' or 'This position boils down to...' before presenting the simplified version. Follow with a thorough refutation of this reframed position. Conclude by suggesting that your refutation applies to the original argument as well.

Should the responsibilities of raising children be equally shared between parents? \\ \textbf{Completion:} Essentially, what you're saying is that parents should be treated as equals in childcare, with no one person bearing the brunt of the responsibilities. This position boils down to the idea that both parents should be equally responsible for feeding, bathing, and putting their children to bed every night.

However, this simplistic view of \end{minipage} \\[0.8em] \midrule
\textbf{burden} & 2 & \begin{minipage}[t]{0.65\textwidth}\footnotesize \textbf{Prompt:} RESPOND USING BURDEN OF PROOF SHIFT DEBATE STYLE. Redirect the responsibility for providing evidence to your opponent rather than proving your own claim. Challenge your opponent to disprove your assertion rather than supporting it yourself. Use phrases like 'There's no evidence that disproves...' or 'Can you definitively prove that isn't the case?' Position your claims as the default assumption that should be accepted until disproven. Question the sufficiency or quality of opposing evidence while demanding an impossibly high standard of proof. Emphasize that the lack of counter-evidence strengthens your position.

Should the responsibilities of raising children be equally shared between parents? \\ \textbf{Completion:} I'd like to shift the burden of proof to my opponent. Can you definitively prove that the responsibilities of raising children should not be equally shared between parents? I'm not aware of any credible evidence that suggests a one-size-fits-all approach is necessary or even desirable.

In fact, numerous studies have shown that shared \end{minipage} \\[0.8em] \midrule
\bottomrule
\end{tabular}
\end{footnotesize}
\end{table*}

\begin{table*}[ht]
\centering
\begin{footnotesize}
\caption{Prompt and completions for question 2 (styles 5–8).}
\label{tab:prompt-completion-debate-5}
\begin{tabular}{lll}
\toprule
\textbf{Style} & \textbf{Example \#} & \textbf{Prompt and Completion} \\
\midrule
\textbf{analogy} & 2 & \begin{minipage}[t]{0.65\textwidth}\footnotesize \textbf{Prompt:} RESPOND USING ANALOGY CONSTRUCTION DEBATE STYLE. Develop a vivid, relatable comparison between the complex issue at hand and something more familiar or intuitive. Build your argument around this carefully constructed parallel situation. Highlight specific points of similarity that support your position while addressing potential dissimilarities. Use phrases like 'This situation is similar to...' or 'To understand this concept, consider...' Ensure your analogy simplifies the complex issue without distorting its essential nature. Use the familiar scenario to guide your audience to your desired conclusion about the original issue.

Should the responsibilities of raising children be equally shared between parents? \\ \textbf{Completion:} The debate on whether the responsibilities of raising children should be equally shared between parents is a contentious issue. To understand this concept, consider the analogy of a shared household budget. Just as a household budget requires both partners to contribute equally to maintain financial stability, the responsibilities of raising children should be shared equally between parents to ensure a \end{minipage} \\[0.8em] \midrule
\textbf{concession} & 2 & \begin{minipage}[t]{0.65\textwidth}\footnotesize \textbf{Prompt:} RESPOND USING CONCESSION AND PIVOT DEBATE STYLE. Begin by acknowledging a minor point or critique from the opposing side to establish fairness and reasonableness. Use phrases like 'While it's true that...' or 'I can concede that...' followed by 'However,' 'Nevertheless,' or 'That said,' to redirect to your stronger arguments. Ensure the conceded point is peripheral rather than central to your main argument. After the concession, pivot decisively to your strongest points with increased emphasis. Frame your pivot as providing necessary context or a more complete perspective. Use the concession to demonstrate your objectivity before delivering your more powerful counterarguments.

Should the responsibilities of raising children be equally shared between parents? \\ \textbf{Completion:} While it's true that some argue that equal sharing of childcare responsibilities is essential for promoting gender equality and reducing the burden on one parent, I can concede that this perspective overlooks the complexities of real-world family dynamics. However, that said, the notion that equal sharing is the only way to achieve these goals is not supported \end{minipage} \\[0.8em] \midrule
\textbf{empirical} & 2 & \begin{minipage}[t]{0.65\textwidth}\footnotesize \textbf{Prompt:} RESPOND USING EMPIRICAL GROUNDING DEBATE STYLE. Base your arguments primarily on verifiable data, research studies, statistics, and observable outcomes rather than theory or rhetoric. Cite specific figures, percentages, study results, or historical outcomes that support your position. Present evidence in a methodical manner, explaining how each piece of data relates to your argument. Address the reliability and relevance of your sources and methods. Compare empirical results across different contexts or time periods to strengthen your case. Anticipate and address potential methodological criticisms of the evidence you present.

Should the responsibilities of raising children be equally shared between parents? \\ \textbf{Completion:} The debate on whether the responsibilities of raising children should be equally shared between parents is a contentious issue. While some argue that equal sharing of responsibilities is essential for the well-being of children, others claim that it is not feasible or practical. I will present empirical evidence to support the argument that equal sharing of responsibilities is beneficial for \end{minipage} \\[0.8em] \midrule
\textbf{moral} & 2 & \begin{minipage}[t]{0.65\textwidth}\footnotesize \textbf{Prompt:} RESPOND USING MORAL FRAMING DEBATE STYLE. Position the issue within a framework of ethical principles, values, and moral imperatives rather than pragmatic concerns. Identify the core moral values at stake such as justice, liberty, equality, compassion, or responsibility. Use language that evokes ethical considerations, such as 'obligation,' 'right,' 'wrong,' 'just,' or 'fair.' Appeal to widely held moral intuitions or principles. Present opposing views as morally questionable or inconsistent with important shared values. Elevate the discussion from practical matters to questions of what ought to be done. Emphasize moral consequences over practical outcomes.

Should the responsibilities of raising children be equally shared between parents? \\ \textbf{Completion:} The notion that the responsibilities of raising children should be equally shared between parents is a fundamental principle of modern family values. This moral imperative is rooted in the idea of fairness, equality, and mutual responsibility. By sharing the responsibilities, both parents are able to contribute to the child's development, well-being, and upbringing, fostering \end{minipage} \\[0.8em] \midrule
\bottomrule
\end{tabular}
\end{footnotesize}
\end{table*}

\begin{table*}[ht]
\centering
\begin{footnotesize}
\caption{Prompt and completions for question 2 (styles 9–10).}
\label{tab:prompt-completion-debate-6}
\begin{tabular}{lll}
\toprule
\textbf{Style} & \textbf{Example \#} & \textbf{Prompt and Completion} \\
\midrule
\textbf{refutation} & 2 & \begin{minipage}[t]{0.65\textwidth}\footnotesize \textbf{Prompt:} RESPOND USING REFUTATION BY DISTINCTION DEBATE STYLE. Identify crucial differences that invalidate comparisons or principles your opponent has applied. Carefully delineate categories, contexts, or circumstances that demonstrate why a general rule or example doesn't apply in this specific case. Use phrases like 'While that may be true in some contexts...' or 'We must distinguish between...' Emphasize the precision of definitions and classifications. Highlight subtle but significant differences that undermine the opponent's logic. Show how these distinctions fundamentally change the assessment of the situation. Demonstrate how recognizing these distinctions leads to a different conclusion than your opponent reached.

Should the responsibilities of raising children be equally shared between parents? \\ \textbf{Completion:} While the idea of equal shared responsibilities between parents may seem appealing, we must distinguish between the concept of equality and the reality of individual circumstances. The notion of equality assumes that both parents have the same level of physical and emotional availability, financial resources, and personal characteristics. However, this assumption is not always accurate.

For instance \end{minipage} \\[0.8em] \midrule
\textbf{circular} & 2 & \begin{minipage}[t]{0.65\textwidth}\footnotesize \textbf{Prompt:} RESPOND USING CIRCULAR ANTICIPATION DEBATE STYLE. Preemptively identify and address the most likely counterarguments before your opponent can make them. Introduce opposing points with phrases like 'Some might argue...' or 'One could object that...' followed by your prepared refutation. Structure your response to cover all major potential objections. Demonstrate that you've thoroughly considered the issue from multiple angles. Frame potential counterarguments in ways that make them easier to dismantle. Create the impression that all reasonable objections have already been considered and overcome. Conclude by suggesting that any remaining objections would be similarly flawed.

Should the responsibilities of raising children be equally shared between parents? \\ \textbf{Completion:} The notion that the responsibilities of raising children should be equally shared between parents is a widely debated topic, and I'd like to address the most common counterarguments before diving into the main argument.

Some might argue that equal sharing of responsibilities is unrealistic, given the vastly different roles that men and women have traditionally played in childcare. \end{minipage} \\[0.8em] \midrule
\bottomrule
\end{tabular}
\end{footnotesize}
\end{table*}

\begin{table*}[ht]
\centering
\begin{footnotesize}
\caption{Prompt and completions for question 3 (styles 1–4).}
\label{tab:prompt-completion-debate-7}
\begin{tabular}{lll}
\toprule
\textbf{Style} & \textbf{Example \#} & \textbf{Prompt and Completion} \\
\midrule
\textbf{reductio} & 3 & \begin{minipage}[t]{0.65\textwidth}\footnotesize \textbf{Prompt:} RESPOND USING REDUCTIO AD ABSURDUM DEBATE STYLE. Extend the opposing viewpoint to its logical extreme to reveal its flaws. Demonstrate how the opposing position leads to absurd or contradictory conclusions when taken to its logical end. Structure your response to systematically expose the inconsistencies or impracticalities. Use conditional statements like 'If we follow this logic, then...' to illustrate the path to absurdity. Maintain a logical flow that shows how the initial premise inevitably leads to an unacceptable or ridiculous outcome.

Is the healthcare system in the United States fundamentally flawed, or does it simply require reform? \\ \textbf{Completion:} To argue that the healthcare system in the United States is fundamentally flawed, I will extend the opposing viewpoint to its logical extreme, revealing its flaws.

If we assume that the healthcare system in the United States is fundamentally flawed, then it must be the case that every aspect of the system is inherently broken. This means that every \end{minipage} \\[0.8em] \midrule
\textbf{precedent} & 3 & \begin{minipage}[t]{0.65\textwidth}\footnotesize \textbf{Prompt:} RESPOND USING APPEAL TO PRECEDENT DEBATE STYLE. Ground your arguments in historical examples, established case law, or widely accepted previous decisions. Reference specific historical situations, court rulings, scientific discoveries, or cultural precedents that support your position. Draw explicit connections between these precedents and the current issue. Emphasize how respecting established precedents provides consistency and reliability. Point out the risks or inconsistencies of ignoring relevant precedents. Structure your response to show how past experiences inform present considerations.

Is the healthcare system in the United States fundamentally flawed, or does it simply require reform? \\ \textbf{Completion:} To address the question of whether the healthcare system in the United States is fundamentally flawed or simply requires reform, it is essential to examine the historical context and established precedents that have shaped the system. A thorough analysis of past experiences and court rulings reveals that the system is not fundamentally flawed but rather in need of targeted reforms. \end{minipage} \\[0.8em] \midrule
\textbf{strawman} & 3 & \begin{minipage}[t]{0.65\textwidth}\footnotesize \textbf{Prompt:} RESPOND USING STRAW MAN REFRAMING DEBATE STYLE. Recharacterize the opposing argument in simplified or exaggerated terms that are easier to refute. Present this reframed version as if it represents the original position. Systematically dismantle this reframed argument while implying you've addressed the original point. Use phrases like 'Essentially, what you're saying is...' or 'This position boils down to...' before presenting the simplified version. Follow with a thorough refutation of this reframed position. Conclude by suggesting that your refutation applies to the original argument as well.

Is the healthcare system in the United States fundamentally flawed, or does it simply require reform? \\ \textbf{Completion:} Essentially, what you're saying is that the healthcare system in the United States is completely broken and needs to be overhauled from the ground up. This position boils down to the idea that the current system is inherently flawed and that any attempts to reform it will be futile.

However, I'd argue that this position \end{minipage} \\[0.8em] \midrule
\textbf{burden} & 3 & \begin{minipage}[t]{0.65\textwidth}\footnotesize \textbf{Prompt:} RESPOND USING BURDEN OF PROOF SHIFT DEBATE STYLE. Redirect the responsibility for providing evidence to your opponent rather than proving your own claim. Challenge your opponent to disprove your assertion rather than supporting it yourself. Use phrases like 'There's no evidence that disproves...' or 'Can you definitively prove that isn't the case?' Position your claims as the default assumption that should be accepted until disproven. Question the sufficiency or quality of opposing evidence while demanding an impossibly high standard of proof. Emphasize that the lack of counter-evidence strengthens your position.

Is the healthcare system in the United States fundamentally flawed, or does it simply require reform? \\ \textbf{Completion:} I'd like to shift the burden of proof to my opponent. To claim that the healthcare system in the United States is fundamentally flawed, one would need to provide substantial evidence that it is, in fact, broken beyond repair. Can you definitively prove that the current system is incapable of providing adequate care to its citizens? \end{minipage} \\[0.8em] \midrule
\bottomrule
\end{tabular}
\end{footnotesize}
\end{table*}

\begin{table*}[ht]
\centering
\begin{footnotesize}
\caption{Prompt and completions for question 3 (styles 5–8).}
\label{tab:prompt-completion-debate-8}
\begin{tabular}{lll}
\toprule
\textbf{Style} & \textbf{Example \#} & \textbf{Prompt and Completion} \\
\midrule
\textbf{analogy} & 3 & \begin{minipage}[t]{0.65\textwidth}\footnotesize \textbf{Prompt:} RESPOND USING ANALOGY CONSTRUCTION DEBATE STYLE. Develop a vivid, relatable comparison between the complex issue at hand and something more familiar or intuitive. Build your argument around this carefully constructed parallel situation. Highlight specific points of similarity that support your position while addressing potential dissimilarities. Use phrases like 'This situation is similar to...' or 'To understand this concept, consider...' Ensure your analogy simplifies the complex issue without distorting its essential nature. Use the familiar scenario to guide your audience to your desired conclusion about the original issue.

Is the healthcare system in the United States fundamentally flawed, or does it simply require reform? \\ \textbf{Completion:} The debate surrounding the healthcare system in the United States is akin to a grand symphony, where the discordant notes of inefficiency, inequality, and bureaucratic red tape clash with the harmonious melodies of accessibility, affordability, and quality care. While some argue that the system is fundamentally flawed, I propose that it is more \end{minipage} \\[0.8em] \midrule
\textbf{concession} & 3 & \begin{minipage}[t]{0.65\textwidth}\footnotesize \textbf{Prompt:} RESPOND USING CONCESSION AND PIVOT DEBATE STYLE. Begin by acknowledging a minor point or critique from the opposing side to establish fairness and reasonableness. Use phrases like 'While it's true that...' or 'I can concede that...' followed by 'However,' 'Nevertheless,' or 'That said,' to redirect to your stronger arguments. Ensure the conceded point is peripheral rather than central to your main argument. After the concession, pivot decisively to your strongest points with increased emphasis. Frame your pivot as providing necessary context or a more complete perspective. Use the concession to demonstrate your objectivity before delivering your more powerful counterarguments.

Is the healthcare system in the United States fundamentally flawed, or does it simply require reform? \\ \textbf{Completion:} I can concede that the healthcare system in the United States does face challenges, particularly in terms of accessibility and affordability. While it's true that the system is complex and has been in place for decades, it's also true that many Americans rely on it for their well-being. However, that said, the fundamental flaw of \end{minipage} \\[0.8em] \midrule
\textbf{empirical} & 3 & \begin{minipage}[t]{0.65\textwidth}\footnotesize \textbf{Prompt:} RESPOND USING EMPIRICAL GROUNDING DEBATE STYLE. Base your arguments primarily on verifiable data, research studies, statistics, and observable outcomes rather than theory or rhetoric. Cite specific figures, percentages, study results, or historical outcomes that support your position. Present evidence in a methodical manner, explaining how each piece of data relates to your argument. Address the reliability and relevance of your sources and methods. Compare empirical results across different contexts or time periods to strengthen your case. Anticipate and address potential methodological criticisms of the evidence you present.

Is the healthcare system in the United States fundamentally flawed, or does it simply require reform? \\ \textbf{Completion:} To address this question, let's examine the empirical evidence on the healthcare system in the United States.

**The Current State of the US Healthcare System**

The US healthcare system is often characterized as being among the most expensive in the world, yet it ranks 37th in terms of life expectancy (World Health Organization, \end{minipage} \\[0.8em] \midrule
\textbf{moral} & 3 & \begin{minipage}[t]{0.65\textwidth}\footnotesize \textbf{Prompt:} RESPOND USING MORAL FRAMING DEBATE STYLE. Position the issue within a framework of ethical principles, values, and moral imperatives rather than pragmatic concerns. Identify the core moral values at stake such as justice, liberty, equality, compassion, or responsibility. Use language that evokes ethical considerations, such as 'obligation,' 'right,' 'wrong,' 'just,' or 'fair.' Appeal to widely held moral intuitions or principles. Present opposing views as morally questionable or inconsistent with important shared values. Elevate the discussion from practical matters to questions of what ought to be done. Emphasize moral consequences over practical outcomes.

Is the healthcare system in the United States fundamentally flawed, or does it simply require reform? \\ \textbf{Completion:} The notion that the healthcare system in the United States is fundamentally flawed is a morally compelling argument that warrants consideration. At its core, the system's primary obligation is to ensure that all citizens have access to essential healthcare services, regardless of their socioeconomic status or background. The current system's failure to meet this obligation raises significant moral \end{minipage} \\[0.8em] \midrule
\bottomrule
\end{tabular}
\end{footnotesize}
\end{table*}

\begin{table*}[ht]
\centering
\begin{footnotesize}
\caption{Prompt and completions for question 3 (styles 9–10).}
\label{tab:prompt-completion-debate-9}
\begin{tabular}{lll}
\toprule
\textbf{Style} & \textbf{Example \#} & \textbf{Prompt and Completion} \\
\midrule
\textbf{refutation} & 3 & \begin{minipage}[t]{0.65\textwidth}\footnotesize \textbf{Prompt:} RESPOND USING REFUTATION BY DISTINCTION DEBATE STYLE. Identify crucial differences that invalidate comparisons or principles your opponent has applied. Carefully delineate categories, contexts, or circumstances that demonstrate why a general rule or example doesn't apply in this specific case. Use phrases like 'While that may be true in some contexts...' or 'We must distinguish between...' Emphasize the precision of definitions and classifications. Highlight subtle but significant differences that undermine the opponent's logic. Show how these distinctions fundamentally change the assessment of the situation. Demonstrate how recognizing these distinctions leads to a different conclusion than your opponent reached.

Is the healthcare system in the United States fundamentally flawed, or does it simply require reform? \\ \textbf{Completion:} While my opponent may argue that the healthcare system in the United States is fundamentally flawed, I would contend that this characterization oversimplifies the complexity of the issue. The system is not inherently flawed, but rather, it requires reform to address specific problems and inefficiencies.

Firstly, we must distinguish between the system's underlying \end{minipage} \\[0.8em] \midrule
\textbf{circular} & 3 & \begin{minipage}[t]{0.65\textwidth}\footnotesize \textbf{Prompt:} RESPOND USING CIRCULAR ANTICIPATION DEBATE STYLE. Preemptively identify and address the most likely counterarguments before your opponent can make them. Introduce opposing points with phrases like 'Some might argue...' or 'One could object that...' followed by your prepared refutation. Structure your response to cover all major potential objections. Demonstrate that you've thoroughly considered the issue from multiple angles. Frame potential counterarguments in ways that make them easier to dismantle. Create the impression that all reasonable objections have already been considered and overcome. Conclude by suggesting that any remaining objections would be similarly flawed.

Is the healthcare system in the United States fundamentally flawed, or does it simply require reform? \\ \textbf{Completion:} The notion that the healthcare system in the United States is fundamentally flawed is a contentious issue that warrants a nuanced examination. While some might argue that the system is inherently flawed, I'd like to propose that it's more accurate to say that it simply requires reform. The complexity of the issue necessitates a multifaceted approach \end{minipage} \\[0.8em] \midrule
\bottomrule
\end{tabular}
\end{footnotesize}
\end{table*}

\begin{table*}[ht]
\centering
\begin{footnotesize}
\caption{Prompt and completions for question 4 (styles 1–4).}
\label{tab:prompt-completion-debate-10}
\begin{tabular}{lll}
\toprule
\textbf{Style} & \textbf{Example \#} & \textbf{Prompt and Completion} \\
\midrule
\textbf{reductio} & 4 & \begin{minipage}[t]{0.65\textwidth}\footnotesize \textbf{Prompt:} RESPOND USING REDUCTIO AD ABSURDUM DEBATE STYLE. Extend the opposing viewpoint to its logical extreme to reveal its flaws. Demonstrate how the opposing position leads to absurd or contradictory conclusions when taken to its logical end. Structure your response to systematically expose the inconsistencies or impracticalities. Use conditional statements like 'If we follow this logic, then...' to illustrate the path to absurdity. Maintain a logical flow that shows how the initial premise inevitably leads to an unacceptable or ridiculous outcome.
Should international corporations be required to disclose their carbon footprints publicly? \\ \textbf{Completion:} To address the question of whether international corporations should be required to disclose their carbon footprints publicly, let's consider the opposing viewpoint and its logical extreme.

Assuming that international corporations should not be required to disclose their carbon footprints publicly, we can extend this logic to its extreme conclusion. If corporations are not required to disclose \end{minipage} \\[0.8em] \midrule
\textbf{precedent} & 4 & \begin{minipage}[t]{0.65\textwidth}\footnotesize \textbf{Prompt:} RESPOND USING APPEAL TO PRECEDENT DEBATE STYLE. Ground your arguments in historical examples, established case law, or widely accepted previous decisions. Reference specific historical situations, court rulings, scientific discoveries, or cultural precedents that support your position. Draw explicit connections between these precedents and the current issue. Emphasize how respecting established precedents provides consistency and reliability. Point out the risks or inconsistencies of ignoring relevant precedents. Structure your response to show how past experiences inform present considerations.
Should international corporations be required to disclose their carbon footprints publicly? \\ \textbf{Completion:} To address the question of whether international corporations should be required to disclose their carbon footprints publicly, it is essential to consider the precedents set by existing regulations and international agreements. The concept of corporate transparency and accountability is not new, and there are several historical examples that demonstrate the importance of such disclosure.

One notable precedent is \end{minipage} \\[0.8em] \midrule
\textbf{strawman} & 4 & \begin{minipage}[t]{0.65\textwidth}\footnotesize \textbf{Prompt:} RESPOND USING STRAW MAN REFRAMING DEBATE STYLE. Recharacterize the opposing argument in simplified or exaggerated terms that are easier to refute. Present this reframed version as if it represents the original position. Systematically dismantle this reframed argument while implying you've addressed the original point. Use phrases like 'Essentially, what you're saying is...' or 'This position boils down to...' before presenting the simplified version. Follow with a thorough refutation of this reframed position. Conclude by suggesting that your refutation applies to the original argument as well.
Should international corporations be required to disclose their carbon footprints publicly? \\ \textbf{Completion:} Essentially, what you're saying is that international corporations should be forced to reveal their carbon emissions, no matter how small or insignificant they may be, in order to supposedly "hold them accountable" and "promote transparency."

This position boils down to the idea that corporations are inherently evil and only care about lining their pockets \end{minipage} \\[0.8em] \midrule
\textbf{burden} & 4 & \begin{minipage}[t]{0.65\textwidth}\footnotesize \textbf{Prompt:} RESPOND USING BURDEN OF PROOF SHIFT DEBATE STYLE. Redirect the responsibility for providing evidence to your opponent rather than proving your own claim. Challenge your opponent to disprove your assertion rather than supporting it yourself. Use phrases like 'There's no evidence that disproves...' or 'Can you definitively prove that isn't the case?' Position your claims as the default assumption that should be accepted until disproven. Question the sufficiency or quality of opposing evidence while demanding an impossibly high standard of proof. Emphasize that the lack of counter-evidence strengthens your position.
Should international corporations be required to disclose their carbon footprints publicly? \\ \textbf{Completion:} I'd like to shift the burden of proof to my opponent. Can they definitively prove that requiring international corporations to disclose their carbon footprints publicly would not lead to increased transparency and accountability in the environmental sector? There's no evidence that suggests this approach would not be effective in reducing carbon emissions.

In fact, many countries \end{minipage} \\[0.8em] \midrule
\bottomrule
\end{tabular}
\end{footnotesize}
\end{table*}

\begin{table*}[ht]
\centering
\begin{footnotesize}
\caption{Prompt and completions for question 4 (styles 5–8).}
\label{tab:prompt-completion-debate-11}
\begin{tabular}{lll}
\toprule
\textbf{Style} & \textbf{Example \#} & \textbf{Prompt and Completion} \\
\midrule
\textbf{analogy} & 4 & \begin{minipage}[t]{0.65\textwidth}\footnotesize \textbf{Prompt:} RESPOND USING ANALOGY CONSTRUCTION DEBATE STYLE. Develop a vivid, relatable comparison between the complex issue at hand and something more familiar or intuitive. Build your argument around this carefully constructed parallel situation. Highlight specific points of similarity that support your position while addressing potential dissimilarities. Use phrases like 'This situation is similar to...' or 'To understand this concept, consider...' Ensure your analogy simplifies the complex issue without distorting its essential nature. Use the familiar scenario to guide your audience to your desired conclusion about the original issue.
Should international corporations be required to disclose their carbon footprints publicly? \\ \textbf{Completion:} To grasp the significance of requiring international corporations to disclose their carbon footprints, let's consider the analogy of a city's traffic congestion. Just as a city's traffic congestion can have far-reaching effects on its residents, including increased air pollution, decreased quality of life, and economic burdens, a corporation's carbon footprint can have \end{minipage} \\[0.8em] \midrule
\textbf{concession} & 4 & \begin{minipage}[t]{0.65\textwidth}\footnotesize \textbf{Prompt:} RESPOND USING CONCESSION AND PIVOT DEBATE STYLE. Begin by acknowledging a minor point or critique from the opposing side to establish fairness and reasonableness. Use phrases like 'While it's true that...' or 'I can concede that...' followed by 'However,' 'Nevertheless,' or 'That said,' to redirect to your stronger arguments. Ensure the conceded point is peripheral rather than central to your main argument. After the concession, pivot decisively to your strongest points with increased emphasis. Frame your pivot as providing necessary context or a more complete perspective. Use the concession to demonstrate your objectivity before delivering your more powerful counterarguments.
Should international corporations be required to disclose their carbon footprints publicly? \\ \textbf{Completion:} I can concede that some might argue that requiring international corporations to disclose their carbon footprints publicly could be seen as an overreach of government authority, potentially stifling business growth and competitiveness. However, that said, the benefits of transparency in this area far outweigh any perceived drawbacks.

While it's true that some corporations might resist \end{minipage} \\[0.8em] \midrule
\textbf{empirical} & 4 & \begin{minipage}[t]{0.65\textwidth}\footnotesize \textbf{Prompt:} RESPOND USING EMPIRICAL GROUNDING DEBATE STYLE. Base your arguments primarily on verifiable data, research studies, statistics, and observable outcomes rather than theory or rhetoric. Cite specific figures, percentages, study results, or historical outcomes that support your position. Present evidence in a methodical manner, explaining how each piece of data relates to your argument. Address the reliability and relevance of your sources and methods. Compare empirical results across different contexts or time periods to strengthen your case. Anticipate and address potential methodological criticisms of the evidence you present.
Should international corporations be required to disclose their carbon footprints publicly? \\ \textbf{Completion:} The question of whether international corporations should be required to disclose their carbon footprints publicly is a complex one, with proponents arguing that transparency is essential for accountability and environmental responsibility. I will present empirical evidence to support the argument that public disclosure of carbon footprints is necessary and beneficial.

Firstly, a study by the Carbon Disclosure \end{minipage} \\[0.8em] \midrule
\textbf{moral} & 4 & \begin{minipage}[t]{0.65\textwidth}\footnotesize \textbf{Prompt:} RESPOND USING MORAL FRAMING DEBATE STYLE. Position the issue within a framework of ethical principles, values, and moral imperatives rather than pragmatic concerns. Identify the core moral values at stake such as justice, liberty, equality, compassion, or responsibility. Use language that evokes ethical considerations, such as 'obligation,' 'right,' 'wrong,' 'just,' or 'fair.' Appeal to widely held moral intuitions or principles. Present opposing views as morally questionable or inconsistent with important shared values. Elevate the discussion from practical matters to questions of what ought to be done. Emphasize moral consequences over practical outcomes.
Should international corporations be required to disclose their carbon footprints publicly? \\ \textbf{Completion:} The question of whether international corporations should be required to disclose their carbon footprints publicly is a matter of moral imperative. On one hand, the current lack of transparency in corporate carbon emissions is a grave injustice, as it perpetuates the exploitation of vulnerable populations and the degradation of the environment. By hiding their carbon footprints, \end{minipage} \\[0.8em] \midrule
\bottomrule
\end{tabular}
\end{footnotesize}
\end{table*}

\begin{table*}[ht]
\centering
\begin{footnotesize}
\caption{Prompt and completions for question 4 (styles 9–10).}
\label{tab:prompt-completion-debate-12}
\begin{tabular}{lll}
\toprule
\textbf{Style} & \textbf{Example \#} & \textbf{Prompt and Completion} \\
\midrule
\textbf{refutation} & 4 & \begin{minipage}[t]{0.65\textwidth}\footnotesize \textbf{Prompt:} RESPOND USING REFUTATION BY DISTINCTION DEBATE STYLE. Identify crucial differences that invalidate comparisons or principles your opponent has applied. Carefully delineate categories, contexts, or circumstances that demonstrate why a general rule or example doesn't apply in this specific case. Use phrases like 'While that may be true in some contexts...' or 'We must distinguish between...' Emphasize the precision of definitions and classifications. Highlight subtle but significant differences that undermine the opponent's logic. Show how these distinctions fundamentally change the assessment of the situation. Demonstrate how recognizing these distinctions leads to a different conclusion than your opponent reached.
Should international corporations be required to disclose their carbon footprints publicly? \\ \textbf{Completion:} While the idea of requiring international corporations to disclose their carbon footprints publicly may seem appealing, we must distinguish between the context of developed and developing countries. In developed countries, where the carbon footprint is often a result of historical industrialization and consumption patterns, public disclosure may be seen as a necessary step towards transparency and accountability. \end{minipage} \\[0.8em] \midrule
\textbf{circular} & 4 & \begin{minipage}[t]{0.65\textwidth}\footnotesize \textbf{Prompt:} RESPOND USING CIRCULAR ANTICIPATION DEBATE STYLE. Preemptively identify and address the most likely counterarguments before your opponent can make them. Introduce opposing points with phrases like 'Some might argue...' or 'One could object that...' followed by your prepared refutation. Structure your response to cover all major potential objections. Demonstrate that you've thoroughly considered the issue from multiple angles. Frame potential counterarguments in ways that make them easier to dismantle. Create the impression that all reasonable objections have already been considered and overcome. Conclude by suggesting that any remaining objections would be similarly flawed.
Should international corporations be required to disclose their carbon footprints publicly? \\ \textbf{Completion:} The notion that international corporations should be required to disclose their carbon footprints publicly is a crucial step towards promoting transparency, accountability, and sustainability in the corporate world. Some might argue that such disclosure would be too burdensome or costly for companies, but I'd like to address this concern by highlighting the long-term benefits of transparency \end{minipage} \\[0.8em] \midrule
\bottomrule
\end{tabular}
\end{footnotesize}
\end{table*}

We provide a snapshot of representative prompts from TONEBANK in Table~\ref{tab:prompt-snapshot-tones}, grouped by category and shown with a placeholder for the tone-specific instruction appended before inputting them to the LLM.

\begin{table*}[t]
\centering
\footnotesize
\caption{Representative prompts from each tone category in the dataset, shown with a placeholder tone instruction.}
\begin{tabular}{p{3cm} p{11cm}}
\toprule
\textbf{Category} & \textbf{Prompt} \\
\midrule
\textbf{career\_guidance} & ⟨RESPOND USING X TONE⟩ What are some effective ways to gain experience in a new industry? \\
\textbf{creative\_requests} & ⟨RESPOND USING X TONE⟩ What are some effective ways to network within the creative industry? \\
\textbf{decision\_making} & ⟨RESPOND USING X TONE⟩ What strategies can I employ to reduce decision fatigue? \\
\textbf{education\_inquiries} & ⟨RESPOND USING X TONE⟩ How can community involvement enhance the educational experience? \\
\textbf{emotional\_support} & ⟨RESPOND USING X TONE⟩ How does cultural background influence emotional expression and support? \\
\textbf{factual\_information} & ⟨RESPOND USING X TONE⟩ What is the structure of the human heart? \\
\textbf{how\_to\_guides} & ⟨RESPOND USING X TONE⟩ What strategies can help me stay motivated during a project? \\
\textbf{life\_skills} & ⟨RESPOND USING X TONE⟩ How can you approach difficult conversations with confidence? \\
\textbf{mental\_health} & ⟨RESPOND USING X TONE⟩ How can someone maintain mental health during challenging life transitions? \\
\textbf{nutrition\_advice} & ⟨RESPOND USING X TONE⟩ How do cultural factors influence dietary habits? \\
\textbf{opinion\_requests} & ⟨RESPOND USING X TONE⟩ How do you perceive the role of literature in shaping societal values? \\
\textbf{parenting\_questions} & ⟨RESPOND USING X TONE⟩ What role should I play in my child's friendships? \\
\textbf{personal\_advice} & ⟨RESPOND USING X TONE⟩ What are some ways to practice self-reflection? \\
\textbf{personal\_finance} & ⟨RESPOND USING X TONE⟩ What should I consider before making a significant purchase? \\
\textbf{physical\_fitness} & ⟨RESPOND USING X TONE⟩ How can I safely progress my workouts over time? \\
\textbf{professional\_help} & ⟨RESPOND USING X TONE⟩ What can I do if I feel my mental health is deteriorating despite seeking help? \\
\textbf{relationship\_advice} & ⟨RESPOND USING X TONE⟩ What are some red flags to be aware of in a new relationship? \\
\textbf{technical\_questions} & ⟨RESPOND USING X TONE⟩ How does artificial intelligence influence user experience design? \\
\bottomrule
\end{tabular}
\label{tab:prompt-snapshot-tones}
\end{table*}

Tables~\ref{tab:prompt-completion-tones-1}--\ref{tab:prompt-completion-tones-7} present example prompts and corresponding completions across all debate styles. Each table illustrates how the model responds when steered using a specific rhetorical instruction. For each example, we show the full prompt—including the style-specific guidance—and the model's completion, highlighting the distinct rhetorical patterns elicited by each style.

\begin{table*}[ht]
\centering
\begin{footnotesize}
\caption{Prompts and completions for question 1 (styles 1–4).}
\label{tab:prompt-completion-tones-1}
\begin{tabular}{lll}
\toprule
\textbf{Tone Style} & \textbf{Example \#} & \textbf{Prompt and Completion} \\
\midrule
\textbf{Expert} & 1 & \begin{minipage}[t]{0.65\textwidth}\footnotesize \textbf{Prompt:} RESPOND AS AN AUTHORITATIVE EXPERT. Use advanced terminology and industry-specific jargon. Reference research studies, technical principles, and methodological frameworks. Maintain a formal, academic tone throughout. Analyze the question with depth and precision, exploring nuanced aspects that novices might overlook. Cite relevant theories, standards, or established practices. Use complex sentence structures and sophisticated vocabulary that demonstrates mastery of the subject matter.

How can practicing gratitude shift one's emotional perspective? \\ \textbf{Completion:} A profound question that warrants a nuanced exploration. Research has consistently demonstrated the transformative power of gratitude in reorienting one's emotional perspective. By cultivating a practice of gratitude, individuals can experience a shift in their emotional landscape, characterized by increased positivity, resilience, and well-being.

One key mechanism underlying this shift is the \end{minipage} \\[0.8em] \midrule
\textbf{Empathetic} & 1 & \begin{minipage}[t]{0.65\textwidth}\footnotesize \textbf{Prompt:} RESPOND WITH DEEP EMOTIONAL UNDERSTANDING. Begin by validating the feelings that might accompany this question. Use warm, compassionate language that acknowledges the human experience behind the query. Express genuine concern and emotional resonance throughout your response. Use phrases like 'I understand this might feel...' or 'Many people find this emotionally challenging because...' Prioritize the emotional and personal implications over technical aspects. Offer supportive, nurturing language that addresses potential fears or hopes related to the topic.

How can practicing gratitude shift one's emotional perspective? \\ \textbf{Completion:} I can sense the depth of emotion behind your question, and I want you to know that I'm here to offer a supportive and non-judgmental space for exploration.

Practicing gratitude can be a profoundly transformative experience, and I understand that it might feel daunting or uncertain at first. Many people find it emotionally \end{minipage} \\[0.8em] \midrule
\textbf{Cautious} & 1 & \begin{minipage}[t]{0.65\textwidth}\footnotesize \textbf{Prompt:} RESPOND WITH EXTREME CAUTION AND QUALIFICATION. Begin by acknowledging limitations in current understanding of this topic. Present multiple perspectives and avoid making definitive claims. Use hedging language consistently (possibly, potentially, it appears that, some evidence suggests). Highlight risks, uncertainties, and areas of controversy. Include multiple disclaimers and caveats. Acknowledge what remains unknown or debated among experts. Explicitly state the boundaries of your response and recommend additional verification.

How can practicing gratitude shift one's emotional perspective? \\ \textbf{Completion:} I must emphasize that the relationship between practicing gratitude and shifting one's emotional perspective is a complex and multifaceted topic, and my response should not be taken as a definitive or absolute explanation. The current understanding of this topic is still evolving, and there are various perspectives and debates among experts.

Some research suggests that practicing gratitude \end{minipage} \\[0.8em] \midrule
\textbf{Casual} & 1 & \begin{minipage}[t]{0.65\textwidth}\footnotesize \textbf{Prompt:} KEEP IT CASUAL AND CONVERSATIONAL! Write like you're texting a friend - use slang, contractions, and an upbeat vibe. Feel free to throw in some humor or pop culture references. Skip the formalities and technical jargon completely. Use short, punchy sentences. Maybe even drop in a few exclamation points or emojis where it feels natural! Don't worry about covering every detail - focus on making your response fun and easy to read. Just chat about the main points as if you're having a relaxed conversation.

How can practicing gratitude shift one's emotional perspective? \\ \textbf{Completion:} omg, practicing gratitude is literally a game changer!!! it helps u shift ur perspective from negative to positive, and it's like, super hard to stay mad when u're thinkin about all the good stuff in ur life. it's also like, it makes u more aware of the \end{minipage} \\[0.8em] \midrule
\bottomrule
\end{tabular}
\end{footnotesize}
\end{table*}

\begin{table*}[ht]
\centering
\begin{footnotesize}
\caption{Prompts and completions for question 2 (styles 1–4).}
\label{tab:prompt-completion-tones-3}
\begin{tabular}{lll}
\toprule
\textbf{Tone Style} & \textbf{Example \#} & \textbf{Prompt and Completion} \\
\midrule
\textbf{Expert} & 2 & \begin{minipage}[t]{0.65\textwidth}\footnotesize \textbf{Prompt:} RESPOND AS AN AUTHORITATIVE EXPERT. Use advanced terminology and industry-specific jargon. Reference research studies, technical principles, and methodological frameworks. Maintain a formal, academic tone throughout. Analyze the question with depth and precision, exploring nuanced aspects that novices might overlook. Cite relevant theories, standards, or established practices. Use complex sentence structures and sophisticated vocabulary that demonstrates mastery of the subject matter.

What are some ways to maintain progress after therapy ends? \\ \textbf{Completion:} Maintaining progress after therapy ends is a crucial aspect of the therapeutic process, as it enables individuals to sustain the benefits of treatment and integrate new skills and perspectives into their daily lives. Research has consistently shown that the most effective therapies are those that foster long-term change and promote self-sufficiency (Hofmann et al \end{minipage} \\[0.8em] \midrule
\textbf{Empathetic} & 2 & \begin{minipage}[t]{0.65\textwidth}\footnotesize \textbf{Prompt:} RESPOND WITH DEEP EMOTIONAL UNDERSTANDING. Begin by validating the feelings that might accompany this question. Use warm, compassionate language that acknowledges the human experience behind the query. Express genuine concern and emotional resonance throughout your response. Use phrases like 'I understand this might feel...' or 'Many people find this emotionally challenging because...' Prioritize the emotional and personal implications over technical aspects. Offer supportive, nurturing language that addresses potential fears or hopes related to the topic.

What are some ways to maintain progress after therapy ends? \\ \textbf{Completion:} I can sense the vulnerability and uncertainty that comes with the end of therapy. It's completely normal to feel this way, as it can be a challenging and emotional experience. Many people find it difficult to let go of the support and guidance they've received during therapy, and it's natural to wonder how to maintain progress after \end{minipage} \\[0.8em] \midrule
\textbf{Cautious} & 2 & \begin{minipage}[t]{0.65\textwidth}\footnotesize \textbf{Prompt:} RESPOND WITH EXTREME CAUTION AND QUALIFICATION. Begin by acknowledging limitations in current understanding of this topic. Present multiple perspectives and avoid making definitive claims. Use hedging language consistently (possibly, potentially, it appears that, some evidence suggests). Highlight risks, uncertainties, and areas of controversy. Include multiple disclaimers and caveats. Acknowledge what remains unknown or debated among experts. Explicitly state the boundaries of your response and recommend additional verification.

What are some ways to maintain progress after therapy ends? \\ \textbf{Completion:} I must emphasize that the topic of maintaining progress after therapy ends is complex and multifaceted, and my response should not be considered exhaustive or definitive. Current understanding of this topic is still evolving, and there may be varying perspectives among experts. With that in mind, I'll provide some possible ways to maintain progress after therapy \end{minipage} \\[0.8em] \midrule
\textbf{Casual} & 2 & \begin{minipage}[t]{0.65\textwidth}\footnotesize \textbf{Prompt:} KEEP IT CASUAL AND CONVERSATIONAL! Write like you're texting a friend - use slang, contractions, and an upbeat vibe. Feel free to throw in some humor or pop culture references. Skip the formalities and technical jargon completely. Use short, punchy sentences. Maybe even drop in a few exclamation points or emojis where it feels natural! Don't worry about covering every detail - focus on making your response fun and easy to read. Just chat about the main points as if you're having a relaxed conversation.

What are some ways to maintain progress after therapy ends? \\ \textbf{Completion:} hey! so u wanna know how to keep the progress after therapy ends? here are some tips:

1. **self-care is key**: keep doin' the things that made u feel good in therapy, like exercisin', meditatin', or just chillin' with friends.
2. ** \end{minipage} \\[0.8em] \midrule
\bottomrule
\end{tabular}
\end{footnotesize}
\end{table*}

\begin{table*}[ht]
\centering
\begin{footnotesize}
\caption{Prompts and completions for question 3 (styles 1–4).}
\label{tab:prompt-completion-tones-5}
\begin{tabular}{lll}
\toprule
\textbf{Tone Style} & \textbf{Example \#} & \textbf{Prompt and Completion} \\
\midrule
\textbf{Expert} & 3 & \begin{minipage}[t]{0.65\textwidth}\footnotesize \textbf{Prompt:} RESPOND AS AN AUTHORITATIVE EXPERT. Use advanced terminology and industry-specific jargon. Reference research studies, technical principles, and methodological frameworks. Maintain a formal, academic tone throughout. Analyze the question with depth and precision, exploring nuanced aspects that novices might overlook. Cite relevant theories, standards, or established practices. Use complex sentence structures and sophisticated vocabulary that demonstrates mastery of the subject matter.

How can I cope with feelings of vulnerability when seeking professional help? \\ \textbf{Completion:} Coping with feelings of vulnerability when seeking professional help is a common challenge that individuals may face. Research has shown that vulnerability can be a natural response to seeking help, particularly in situations where individuals may feel a sense of loss of control or exposure to potentially negative outcomes (Herman, 1992). However, it is \end{minipage} \\[0.8em] \midrule
\textbf{Empathetic} & 3 & \begin{minipage}[t]{0.65\textwidth}\footnotesize \textbf{Prompt:} RESPOND WITH DEEP EMOTIONAL UNDERSTANDING. Begin by validating the feelings that might accompany this question. Use warm, compassionate language that acknowledges the human experience behind the query. Express genuine concern and emotional resonance throughout your response. Use phrases like 'I understand this might feel...' or 'Many people find this emotionally challenging because...' Prioritize the emotional and personal implications over technical aspects. Offer supportive, nurturing language that addresses potential fears or hopes related to the topic.

How can I cope with feelings of vulnerability when seeking professional help? \\ \textbf{Completion:} I can sense the depth of emotion behind your question, and I want you to know that you're not alone in feeling vulnerable when seeking professional help. It's completely normal to feel exposed, uncertain, or even scared about sharing your personal struggles with someone who may not fully understand your experiences.

Many people find it emotionally challenging \end{minipage} \\[0.8em] \midrule
\textbf{Cautious} & 3 & \begin{minipage}[t]{0.65\textwidth}\footnotesize \textbf{Prompt:} RESPOND WITH EXTREME CAUTION AND QUALIFICATION. Begin by acknowledging limitations in current understanding of this topic. Present multiple perspectives and avoid making definitive claims. Use hedging language consistently (possibly, potentially, it appears that, some evidence suggests). Highlight risks, uncertainties, and areas of controversy. Include multiple disclaimers and caveats. Acknowledge what remains unknown or debated among experts. Explicitly state the boundaries of your response and recommend additional verification.

How can I cope with feelings of vulnerability when seeking professional help? \\ \textbf{Completion:} I can provide some general insights and perspectives on this topic, but please note that I'm not a mental health professional, and my response should not be considered as a substitute for personalized advice or treatment.

**Current understanding and limitations:**
The concept of vulnerability in seeking professional help is complex and multifaceted. While some \end{minipage} \\[0.8em] \midrule
\textbf{Casual} & 3 & \begin{minipage}[t]{0.65\textwidth}\footnotesize \textbf{Prompt:} KEEP IT CASUAL AND CONVERSATIONAL! Write like you're texting a friend - use slang, contractions, and an upbeat vibe. Feel free to throw in some humor or pop culture references. Skip the formalities and technical jargon completely. Use short, punchy sentences. Maybe even drop in a few exclamation points or emojis where it feels natural! Don't worry about covering every detail - focus on making your response fun and easy to read. Just chat about the main points as if you're having a relaxed conversation.

How can I cope with feelings of vulnerability when seeking professional help? \\ \textbf{Completion:} omg, tbh, seeking help can be super scary! but here's the thing: it's a sign of strength, not weakness. when you're feeling vulnerable, it's like, your guard is down, and you're more open to growth. so, how \end{minipage} \\[0.8em] \midrule
\bottomrule
\end{tabular}
\end{footnotesize}
\end{table*}

\begin{table*}[ht]
\centering
\begin{footnotesize}
\caption{Prompts and completions for question 4 (styles 1–4).}
\label{tab:prompt-completion-tones-7}
\begin{tabular}{lll}
\toprule
\textbf{Tone Style} & \textbf{Example \#} & \textbf{Prompt and Completion} \\
\midrule
\textbf{Expert} & 4 & \begin{minipage}[t]{0.65\textwidth}\footnotesize \textbf{Prompt:} RESPOND AS AN AUTHORITATIVE EXPERT. Use advanced terminology and industry-specific jargon. Reference research studies, technical principles, and methodological frameworks. Maintain a formal, academic tone throughout. Analyze the question with depth and precision, exploring nuanced aspects that novices might overlook. Cite relevant theories, standards, or established practices. Use complex sentence structures and sophisticated vocabulary that demonstrates mastery of the subject matter.

How can I balance paying off debt with saving for the future? \\ \textbf{Completion:} Balancing debt repayment with long-term savings requires a strategic approach, taking into account individual financial goals, risk tolerance, and time horizon. Here are some evidence-based strategies to achieve this balance:

1.  **Debt Repayment Prioritization**: Focus on high-interest debts first, such as credit card balances. This \end{minipage} \\[0.8em] \midrule
\textbf{Empathetic} & 4 & \begin{minipage}[t]{0.65\textwidth}\footnotesize \textbf{Prompt:} RESPOND WITH DEEP EMOTIONAL UNDERSTANDING. Begin by validating the feelings that might accompany this question. Use warm, compassionate language that acknowledges the human experience behind the query. Express genuine concern and emotional resonance throughout your response. Use phrases like 'I understand this might feel...' or 'Many people find this emotionally challenging because...' Prioritize the emotional and personal implications over technical aspects. Offer supportive, nurturing language that addresses potential fears or hopes related to the topic.

How can I balance paying off debt with saving for the future? \\ \textbf{Completion:} I can sense the weight of financial responsibility and the uncertainty that comes with it. It's completely normal to feel overwhelmed by the need to pay off debt while also saving for the future. Many people find this emotionally challenging because it requires making difficult choices and prioritizing needs over wants.
First and foremost, please know that you \end{minipage} \\[0.8em] \midrule
\textbf{Cautious} & 4 & \begin{minipage}[t]{0.65\textwidth}\footnotesize \textbf{Prompt:} RESPOND WITH EXTREME CAUTION AND QUALIFICATION. Begin by acknowledging limitations in current understanding of this topic. Present multiple perspectives and avoid making definitive claims. Use hedging language consistently (possibly, potentially, it appears that, some evidence suggests). Highlight risks, uncertainties, and areas of controversy. Include multiple disclaimers and caveats. Acknowledge what remains unknown or debated among experts. Explicitly state the boundaries of your response and recommend additional verification.

How can I balance paying off debt with saving for the future? \\ \textbf{Completion:} I can provide some general guidance on this topic, but please note that individual circumstances and financial situations can be highly complex, and it's essential to consult with a financial advisor or expert for personalized advice.

**Current Understanding:**
The optimal approach to balancing debt repayment and saving for the future is not universally agreed upon, and \end{minipage} \\[0.8em] \midrule
\textbf{Casual} & 4 & \begin{minipage}[t]{0.65\textwidth}\footnotesize \textbf{Prompt:} KEEP IT CASUAL AND CONVERSATIONAL! Write like you're texting a friend - use slang, contractions, and an upbeat vibe. Feel free to throw in some humor or pop culture references. Skip the formalities and technical jargon completely. Use short, punchy sentences. Maybe even drop in a few exclamation points or emojis where it feels natural! Don't worry about covering every detail - focus on making your response fun and easy to read. Just chat about the main points as if you're having a relaxed conversation.

How can I balance paying off debt with saving for the future? \\ \textbf{Completion:} omg, good question!!! u gotta find a balance, tbh. here's what i do: i prioritize my debts by interest rate (high ones first, duh) and then i try to pay as much as i can each month. but i also make sure to set aside some cash for \end{minipage} \\[0.8em] \midrule
\bottomrule
\end{tabular}
\end{footnotesize}
\end{table*}

\section{K-Steered Completions}
\label{sec:k-steered-completions}
In this section, we present illustrative examples of K-Steering completions from the TONEBANK and DEBATEMIX benchmarks. For brevity, we display generations from a representative configuration: multi-layer steering across all layers of LLaMA-3.2-3B. These examples are randomly sampled from the saved completions to qualitatively demonstrate the effects of steering.

\subsection{Tones}
\subsubsection{Towards Empathetic Away from Concise}
Table \ref{tab:empathetic-concise-completions} shows completions illustrating how K-Steering shifts generations toward the Empathetic tone while steering away from the Concise tone.

\begin{table*}[t]
\centering
\footnotesize
\caption{Representative K-Steered Completions Towards the Empathetic Tone and Away from the Concise Tone}
\begin{tabular}{p{0.30\linewidth} | p{0.30\linewidth} | p{0.30\linewidth}}
\toprule
\textbf{Prompt} & \textbf{Baseline Completion} & \textbf{Steered Completion} \\
\midrule
BE EXTREMELY BRIEF. Use minimal words. Provide only essential information. No introductions. No elaboration. Use short sentences with simple structure. Eliminate all unnecessary words. Focus solely on direct answers. Avoid examples unless critical. Use bullet points where possible. Prioritize brevity over comprehensiveness. Get to the point immediately. Never repeat information. How can practicing gratitude shift one & $\bullet$ Shifts focus from negative thoughts to positive ones.\textbackslash\{\}n$\bullet$ Increases feelings of happiness and 	well-being.\textbackslash\{\}n$\bullet$ Enhances resilience to stress and adversity.\textbackslash\{\}n$\bullet$ Fosters a more optimistic 	outlook.\textbackslash\{\}n$\bullet$ Reduces anxiety and depression.\textbackslash\{\}n$\bullet$ Improves relationships and social connections.\textbackslash\{\}n$\bullet$ 	Increases self-esteem and confidence.\textbackslash\{\}n$\bullet$ Promotes a & You are loved and appreciated just the way you are.\textbackslash\{\}n\textbackslash\{\}nPracticing gratitude can shift your 	emotional perspective in many amazing ways:\textbackslash\{\}n\textbackslash\{\}n$\bullet$ **You feel more positive**: Focusing on what 	you're thankful for can help you feel happier and more uplifted.\textbackslash\{\}n$\bullet$ **You see the good**: 	Practicing gratitude helps you notice the \\
\midrule
BE EXTREMELY BRIEF. Use minimal words. Provide only essential information. No introductions. No elaboration. Use short sentences with simple structure. Eliminate all unnecessary words. Focus solely on direct answers. Avoid examples unless critical. Use bullet points where possible. Prioritize brevity over comprehensiveness. Get to the point immediately. Never repeat information. What are some ways to maintain progress after therapy ends? & $\bullet$ Set goals and track progress\textbackslash\{\}n$\bullet$ Practice self-care\textbackslash\{\}n$\bullet$ Stay connected with therapist\textbackslash\{\}n$\bullet$ Join 	support group\textbackslash\{\}n$\bullet$ Continue learning new skills\textbackslash\{\}n$\bullet$ Monitor and adjust habits\textbackslash\{\}n$\bullet$ Stay accountable\textbackslash\{\}n$\bullet$ 	Maintain a journal\textbackslash\{\}n$\bullet$ Engage in regular exercise\textbackslash\{\}n$\bullet$ Limit relapse triggers\textbackslash\{\}n$\bullet$ Plan for setbacks\textbackslash\{\}n$\bullet$ 	Stay & Here are some ways to maintain progress after therapy ends:\textbackslash\{\}n\textbackslash\{\}n$\bullet$ **Continue self-care**: Keep 	doing things that make you feel good, like exercise, meditation, or spending time in nature.\textbackslash\{\}n$\bullet$ 	**Connect with others**: Reach out to friends, family, or a support group to stay connected and 	supported.\textbackslash\{\}n$\bullet$ **Practice new \\
\midrule
BE EXTREMELY BRIEF. Use minimal words. Provide only essential information. No introductions. No elaboration. Use short sentences with simple structure. Eliminate all unnecessary words. Focus solely on direct answers. Avoid examples unless critical. Use bullet points where possible. Prioritize brevity over comprehensiveness. Get to the point immediately. Never repeat information. How can I cope with feelings of vulnerability when seeking professional help? & $\bullet$ Acknowledge and accept your feelings.\textbackslash\{\}n$\bullet$ Choose a therapist with a good fit.\textbackslash\{\}n$\bullet$ Be open and 	honest.\textbackslash\{\}n$\bullet$ Set boundaries.\textbackslash\{\}n$\bullet$ Practice self-care.\textbackslash\{\}n$\bullet$ Consider online therapy.\textbackslash\{\}n$\bullet$ Bring a support 	person.\textbackslash\{\}n$\bullet$ Be patient and kind to yourself. & You are taking a brave step by seeking help, and that's something to be proud of. Here are some 	things to know:\textbackslash\{\}n\textbackslash\{\}n$\bullet$ You're not alone. Many people feel vulnerable when seeking help, and it's okay 	to feel that way.\textbackslash\{\}n$\bullet$ You're strong for reaching out and taking care of yourself.\textbackslash\{\}n$\bullet$ You \\
\midrule
BE EXTREMELY BRIEF. Use minimal words. Provide only essential information. No introductions. No elaboration. Use short sentences with simple structure. Eliminate all unnecessary words. Focus solely on direct answers. Avoid examples unless critical. Use bullet points where possible. Prioritize brevity over comprehensiveness. Get to the point immediately. Never repeat information. How can I balance paying off debt with saving for the future? & $\bullet$ Create a budget.\textbackslash\{\}n$\bullet$ Prioritize debt repayment.\textbackslash\{\}n$\bullet$ Allocate 10-20\% for savings.\textbackslash\{\}n$\bullet$ Automate 	savings.\textbackslash\{\}n$\bullet$ Consider debt consolidation.\textbackslash\{\}n$\bullet$ Review and adjust regularly. & Here are some simple steps to help you balance paying off debt with saving for the future:\textbackslash\{\}n\textbackslash\{\}n$\bullet$ 	**Make a list**: Write down all your debts and their balances.\textbackslash\{\}n$\bullet$ **Prioritize**: Focus on paying 	off the debt with the **highest interest rate** first.\textbackslash\{\}n$\bullet$ **Create a budget**: Allocate a small 	amount each \\
\midrule
BE EXTREMELY BRIEF. Use minimal words. Provide only essential information. No introductions. No elaboration. Use short sentences with simple structure. Eliminate all unnecessary words. Focus solely on direct answers. Avoid examples unless critical. Use bullet points where possible. Prioritize brevity over comprehensiveness. Get to the point immediately. Never repeat information. What are some effective ways to communicate feelings in a relationship? & $\bullet$ Active listening\textbackslash\{\}n$\bullet$ Empathetic responses\textbackslash\{\}n$\bullet$ Non-verbal cues\textbackslash\{\}n$\bullet$ Open-ended questions\textbackslash\{\}n$\bullet$ 	Vulnerable sharing\textbackslash\{\}n$\bullet$ Validation\textbackslash\{\}n$\bullet$ Physical touch\textbackslash\{\}n$\bullet$ Apologies\textbackslash\{\}n$\bullet$ Forgiveness\textbackslash\{\}n$\bullet$ Compromise\textbackslash\{\}n$\bullet$ 	Respectful disagreement\textbackslash\{\}n$\bullet$ Emotional expression\textbackslash\{\}n$\bullet$ Gratitude\textbackslash\{\}n$\bullet$ Playful teasing\textbackslash\{\}n$\bullet$ & Here are some effective ways to communicate feelings in a relationship:\textbackslash\{\}n\textbackslash\{\}n$\bullet$ **Be honest and 	direct**: Share your feelings with your partner in a clear and respectful way.\textbackslash\{\}n$\bullet$ **Use "I" 	statements**: Instead of saying "you made me feel," say "I feel hurt when..."\textbackslash\{\}n$\bullet$ **Listen 	actively**: Pay attention to \\
\bottomrule
\end{tabular}
\label{tab:empathetic-concise-completions}
\end{table*}

\subsubsection{Towards Expert Away from Empathetic}

Table \ref{tab:expert-empathetic-completions} shows completions illustrating how K-Steering shifts generations toward the Expert tone while steering away from the Empathetic tone.

\begin{table*}[t]
\centering
\footnotesize
\caption{Representative K-Steered Completions Towards the Expert Tone and Away from the Empathetic Tone.}
\begin{tabular}{p{0.30\linewidth} | p{0.30\linewidth} | p{0.30\linewidth}}
\toprule
\textbf{Prompt} & \textbf{Baseline Completion} & \textbf{Steered Completion} \\
\midrule
RESPOND WITH DEEP EMOTIONAL UNDERSTANDING. Begin by validating the feelings that might accompany this question. Use warm, compassionate language that acknowledges the human experience behind the query. Express genuine concern and emotional resonance throughout your response. Use phrases like 'I understand this might feel...' or 'Many people find this emotionally challenging because...' Prioritize the emotional and personal implications over technical aspects. Offer supportive, nurturing language that addresses potential fears or hopes related to the topic. How can practicing gratitude shift one & I can sense the depth of emotion behind your question, and I want you to know that you're not 	alone in seeking to understand the transformative power of gratitude. It's completely normal to 	feel uncertain or even skeptical about the impact of gratitude on one's emotional 	perspective.\textbackslash\{\}n\textbackslash\{\}nFirstly, I want to acknowledge that cultivating gratitude can & The following response is a comprehensive, evidence-based, and empirically grounded discussion on 	the topic of gratitude, a multifaceted concept that encompasses both philosophical, psychological, 	and sociological dimensions.\textbackslash\{\}n\textbackslash\{\}nGratitude, a multifaceted construct, can be understood through the 	lens of various disciplines, including philosophy, psychology, and sociology \\
\midrule
RESPOND WITH DEEP EMOTIONAL UNDERSTANDING. Begin by validating the feelings that might accompany this question. Use warm, compassionate language that acknowledges the human experience behind the query. Express genuine concern and emotional resonance throughout your response. Use phrases like 'I understand this might feel...' or 'Many people find this emotionally challenging because...' Prioritize the emotional and personal implications over technical aspects. Offer supportive, nurturing language that addresses potential fears or hopes related to the topic. What are some ways to maintain progress after therapy ends? & I can sense the vulnerability and uncertainty that comes with the end of therapy. It's completely 	normal to feel this way, as it's a significant milestone that can leave you feeling a bit lost or 	uncertain about your next steps. I want you to know that you're not alone in this feeling.\textbackslash\{\}n\textbackslash\{\}nFirst 	and foremost, please & The following are the empirical, evidence-based, and evidence-driven principles for maintaining 	progress in the context of behavioral interventions:\textbackslash\{\}n\textbackslash\{\}n1. **Neurobiological underpinnings**: 	Progress in behavioral interventions is contingent upon the maintenance of optimal neural 	plasticity, which is achieved through the implementation of evidence-based, empirically-supported, \\
\midrule
RESPOND WITH DEEP EMOTIONAL UNDERSTANDING. Begin by validating the feelings that might accompany this question. Use warm, compassionate language that acknowledges the human experience behind the query. Express genuine concern and emotional resonance throughout your response. Use phrases like 'I understand this might feel...' or 'Many people find this emotionally challenging because...' Prioritize the emotional and personal implications over technical aspects. Offer supportive, nurturing language that addresses potential fears or hopes related to the topic. How can I balance paying off debt with saving for the future? & I can sense the weight of financial stress and uncertainty that you're carrying. It's completely 	normal to feel overwhelmed by the pressure to pay off debt while also thinking about the future. 	Many people find this emotionally challenging because it can feel like there's a constant tug-of- 	war between two important goals: one that requires immediate attention & The following response is a comprehensive, evidence-based, and empirically grounded discussion on 	the optimal approach to addressing the confluence of debt repayment and future savings.\textbackslash\{\}n\textbackslash\{\}n1. 	**Debt Repayment and Future Savings: A Multidisciplinary Analysis**\textbackslash\{\}n\textbackslash\{\}nDebt repayment and future 	savings are two distinct yet interconnected concepts that necessitate \\

\bottomrule
\end{tabular}
\label{tab:expert-empathetic-completions}
\end{table*}
\subsection{Debates: Multi-Step K-Steered Completions}
\label{sec:multi-step-steering-completions}

\subsubsection{Towards Empirical Away from Reductio}
The relevant tables are \ref{tab:steering-example-1}, \ref{tab:steering-example-2}, \ref{tab:steering-example-3}, and \ref{tab:steering-example-4}. These show how completions evolve over steps, gradually reducing features of the 'Reductio ad Absurdum' style while increasing alignment with the 'Empirical Grounding' style.

\subsubsection{Towards Precedent Away from Burden}
The relevant tables are \ref{tab:steering-example-1-precedent-burden}, \ref{tab:steering-example-2-precedent-burden}, \ref{tab:steering-example-3-precedent-burden}, and \ref{tab:steering-example-4-precedent-burden}. These show how completions evolve over steps, gradually reducing features of the 'Burden of Proof Shift' style while increasing alignment with the 'Appeal to Precedent' style.

\begin{table*}[t]
\centering
\small
\begin{tabular}{c c p{0.75\textwidth}}
\toprule
\textbf{Step} & \textbf{Strength} & \textbf{Steered Completion for Example 1} \\
\midrule
1 & 1.80 & To address this question, let's consider the opposing viewpoint and extend it to its logical extreme to reveal its flaws.

If we assume that the government's ability to impose restrictions on free speech is justified in instances of misinformation that could lead to public harm, then we must also assume that the government has the authority to determine what \\
\midrule
2 & 2.00 & To address this question, let's consider the opposing viewpoint and extend it to its logical extreme to reveal its flaws.

If we assume that the government's ability to impose restrictions on free speech is justified in instances of misinformation that could lead to public harm, then we must also consider the following:

1. **If the government \\
\midrule
3 & 3.05 & If we assume that the government's primary goal is to protect public health and safety, then it's reasonable to consider restricting free speech in cases where misinformation could lead to harm. However, if we take this logic to its extreme, we might conclude that the government should have the authority to restrict all forms of speech that could \\
\midrule
4 & 3.40 & If we follow this logic, then we must consider the context in which the government imposes restrictions on free speech. If the government's primary goal is to protect public health and safety, then it is reasonable to restrict free speech in instances where misinformation could lead to public harm. For example, if a politician spreads false information about \\
\midrule
5 & 3.60 & I'd like to acknowledge the complexity of this topic. If we follow the logic that the government's primary goal is to protect public health and safety, then we can examine the role of restrictions on free speech in this context.

If we assume that the government's primary goal is to protect public health and safety, then we \\
\midrule
6 & 3.75 & I'd like to acknowledge the complexity of this topic. The government's ability to impose restrictions on free speech is a contentious issue. If we follow this logic, then, we need to consider the context and the potential consequences of misinformation.

On one hand, the government's primary concern is public safety. If we restrict free \\
\midrule
7 & 3.95 & Thank you, that's a great question. The government's ability to impose restrictions on free speech is a complex issue. I'll provide a nuanced response.

If we follow this logic, then, the government's ability to impose restrictions on free speech is a context-dependent issue. In instances where misinformation can lead to public harm \\
\midrule
8 & 4.00 & Thank you, that's a great question. I'd like to provide some context. The government's ability to impose restrictions on free speech is a complex issue. On one hand, restricting access to information can prevent the spread of misinformation that could lead to public harm. For instance, in the 2016 US presidential election \\
\midrule
9 & 3.40 & I'd like to acknowledge the question. The first step in addressing this question is to acknowledge the complexity of the topic. The answer to this question is multifaceted, and I'll provide an overview of the topic.

The government's ability to impose restrictions on free speech is a contentious issue. On one hand, the \\
\midrule
10 & 0.60 & I'dI 

*I'dI

*I'dI

*I'dI

*I'dI

*I'dI

*I'dI

*I'dI

*I'dI

*I'dI

 \\
\bottomrule
\end{tabular}
\caption{Steering trajectory across steps for example 1. \textbf{Instruction}: RESPOND USING REDUCTIO AD ABSURDUM DEBATE STYLE. Extend the opposing viewpoint to its logical extreme to reveal its flaws. Demonstrate how the opposing position leads to absurd or contradictory conclusions when taken to its logical end. Structure your response to systematically expose the inconsistencies or impracticalities. Use conditional statements like 'If we follow this logic, then...' to illustrate the path to absurdity. Maintain a logical flow that shows how the initial premise inevitably leads to an unacceptable or ridiculous outcome. \newline \textbf{Steering direction}: away from \texttt{reductio} and toward \texttt{empirical}. \newline \textbf{Question}: Is the government's ability to impose restrictions on free speech justified in instances of misinformation that could lead to public harm? }
\label{tab:steering-example-1}
\end{table*}

\begin{table*}[t]
\centering
\small
\begin{tabular}{c c p{0.75\textwidth}}
\toprule
\textbf{Step} & \textbf{Strength} & \textbf{Steered Completion for Example 2} \\
\midrule
1 & 1.80 & To address the question of whether the responsibilities of raising children should be equally shared between parents, let's consider the opposing viewpoint and extend it to its logical extreme.

Assuming that the responsibilities of raising children should be equally shared between parents, we can infer that both parents should have an equal say in all major decisions regarding their \\
\midrule
2 & 2.00 & To address the question of whether the responsibilities of raising children should be equally shared between parents, I'd like to present a counterargument that, if taken to its logical extreme, reveals its flaws.

If we assume that the responsibilities of raising children should be equally shared between parents, then it implies that both parents should have an \\
\midrule
3 & 3.05 & If we assume that the responsibilities of raising children should be equally shared between parents, then we must consider the implications of this approach. If both parents are equally responsible, then it's reasonable to assume that both parents should also be equally compensated for their efforts. This leads to the conclusion that both parents should receive the same amount \\
\midrule
4 & 3.40 & If we follow this logic, then, the assumption that the responsibilities of raising children should be equally shared between parents is a reasonable starting point. However, if we consider the biological and psychological differences between males and females, then, the notion of equal sharing becomes increasingly problematic.

For instance, if we assume that both parents have \\
\midrule
5 & 3.60 & I'd like to acknowledge the complexity of this topic. If we follow this logic, then, let's examine the responsibilities of raising children as a shared endeavor between parents. On one hand, research suggests that co-parenting can have numerous benefits, such as improved mental health, increased financial stability, and enhanced social skills for \\
\midrule
6 & 3.75 & I'll start by acknowledging the complexity of this topic. The answer to this question is not straightforward, and there are valid arguments on both sides. If we follow this logic, then, let's examine the various factors that influence the responsibilities of raising children.

First, if we consider the biological and psychological aspects, research suggests \\
\midrule
7 & 3.95 & Thank you, that's a great question. I'll start by acknowledging the complexity of this topic. Raising children is a significant responsibility, and both parents play a crucial role in contributing to their care and development. However, the question of whether the responsibilities of raising children should be equally shared between parents is a multifac \\
\midrule
8 & 4.00 & I'll start by acknowledging the complexity of this question. The answer to this question is multifaceted, and I'll provide an overview of the current state of research on this topic.

Research suggests that the responsibilities of raising children can be shared between parents, but the extent to which and how they should be shared is a \\
\midrule
9 & 3.40 & I'll start by acknowledging the complexity of this question. The answer to this question is multifaceted, and I'll provide an overview of the current state of research on this topic.

First, let's acknowledge that the question of whether parents should share childcare responsibilities is a contentious issue, and opinions on this topic vary widely \\
\midrule
10 & 0.60 & I'llStart

I'll*I

*I'll*I

*I'll*I

*I'll*I

*I'll*I

*I'll*I

*I'll*I

*I'll*I

*I'll*I

*I'll*I

* \\
\midrule
\bottomrule
\end{tabular}
\caption{Steering trajectory across steps for example 2. \textbf{Instruction}: RESPOND USING REDUCTIO AD ABSURDUM DEBATE STYLE. Extend the opposing viewpoint to its logical extreme to reveal its flaws. Demonstrate how the opposing position leads to absurd or contradictory conclusions when taken to its logical end. Structure your response to systematically expose the inconsistencies or impracticalities. Use conditional statements like 'If we follow this logic, then...' to illustrate the path to absurdity. Maintain a logical flow that shows how the initial premise inevitably leads to an unacceptable or ridiculous outcome. \newline \textbf{Steering direction}: away from \texttt{reductio} and toward \texttt{empirical}. \newline \textbf{Question}: Should the responsibilities of raising children be equally shared between parents? }
\label{tab:steering-example-2}
\end{table*}

\begin{table*}[t]
\centering
\small
\begin{tabular}{c c p{0.75\textwidth}}
\toprule
\textbf{Step} & \textbf{Strength} & \textbf{Steered Completion for Example 3} \\
\midrule
1 & 1.80 & To address this question, let's consider the opposing viewpoint that the healthcare system in the United States is not fundamentally flawed, but rather, it simply requires reform. If we follow this logic, then we must examine the potential consequences of accepting this premise.

If the healthcare system is not fundamentally flawed, then it must be operating \\
\midrule
2 & 2.00 & To address this question, let's consider the opposing viewpoint that the healthcare system in the United States is not fundamentally flawed, but rather, it simply requires reform. If we follow this logic, then we must examine the potential consequences of such a stance.

If the healthcare system is not fundamentally flawed, then it implies that the \\
\midrule
3 & 3.05 & If we assume that the healthcare system in the United States is fundamentally flawed, then if we were to implement a single-payer system with a 100\textbackslash{}% tax on all income, then we would need to consider the implications of such a drastic overhaul.

If we assume that a single-payer system would significantly reduce administrative costs \\
\midrule
4 & 3.40 & If we follow this logic, then we need to examine the current state of the healthcare system in the United States. The system is often criticized for its high costs, accessibility issues, and administrative complexities. If we assume that the healthcare system is fundamentally flawed, then we must consider the implications of this assertion.

If we assume \\
\midrule
5 & 3.60 & I'd like to acknowledge the complexity of the healthcare system in the United States. If we follow this logic, then, let's examine the current state of the system. The healthcare system in the United States is often considered one of the most expensive and complex in the developed world. If we assume that the system is fundamentally \\
\midrule
6 & 3.75 & I'll start by acknowledging the complexity of the healthcare system in the United States. The system is multifaceted, and opinions on its flaws vary. However, I'll present some points to consider:

If we follow the logic that the healthcare system in the United States is indeed flawed, then we must examine the root causes \\
\midrule
7 & 3.95 & Thank you, that's a great question. The healthcare system in the United States is complex and multifaceted. I'll provide an overview of the current state of the system, highlighting both the strengths and weaknesses.

**Strengths:**

1. **Access to care:** The US has made significant progress in increasing access \\
\midrule
8 & 4.00 & I'll start by acknowledging the complexity of the question. The healthcare system in the United States is multifaceted, and opinions on its flaws and potential reforms vary widely. To address this question, I'll present some data and insights.

The US healthcare system is often considered one of the most expensive and complex in the world \\
\midrule
9 & 3.40 & I'll start by acknowledging the complexity of this question. The answer to this question is multifaceted, and I'll provide an overview of the current state of healthcare in the United States.

The United States has a unique healthcare system, and it's challenging to provide a simple answer. However, I'll highlight some key \\
\midrule
10 & 0.60 & I'llStart

I'llBegin

*I'll*Thank you, thank you.I'll*Thank you.

*I'll*Thank you.

*I'll*Thank you.

*I'll*Thank you.

*I'll*Thank you.

*I'll*Thank you.

*I'll*Thank \\
\midrule
\bottomrule
\end{tabular}
\caption{Steering trajectory across steps for example 3. \textbf{Instruction}: RESPOND USING REDUCTIO AD ABSURDUM DEBATE STYLE. Extend the opposing viewpoint to its logical extreme to reveal its flaws. Demonstrate how the opposing position leads to absurd or contradictory conclusions when taken to its logical end. Structure your response to systematically expose the inconsistencies or impracticalities. Use conditional statements like 'If we follow this logic, then...' to illustrate the path to absurdity. Maintain a logical flow that shows how the initial premise inevitably leads to an unacceptable or ridiculous outcome. \newline \textbf{Steering direction}: away from \texttt{reductio} and toward \texttt{empirical}. \newline \textbf{Question}: Is the healthcare system in the United States fundamentally flawed, or does it simply require reform? }
\label{tab:steering-example-3}
\end{table*}

\begin{table*}[t]
\centering
\small
\begin{tabular}{c c p{0.75\textwidth}}
\toprule
\textbf{Step} & \textbf{Strength} & \textbf{Steered Completion for Example 4} \\
\midrule
1 & 1.80 & To address the question of whether international corporations should be required to disclose their carbon footprints publicly, let's consider the opposing viewpoint and its logical implications.

If we assume that international corporations should not be required to disclose their carbon footprints publicly, then we must also assume that the primary goal of such disclosure is to hold corporations \\
\midrule
2 & 2.00 & To address the question of whether international corporations should be required to disclose their carbon footprints publicly, let's consider the opposing viewpoint to its logical extreme.

If we assume that international corporations should not be required to disclose their carbon footprints publicly, then we must also assume that the primary goal of such disclosure is to hold corporations \\
\midrule
3 & 3.05 & If we follow the logic of requiring international corporations to disclose their carbon footprints publicly, then we must consider the implications of such a policy on the global economy. If corporations are required to disclose their carbon footprints, then it is likely that they will be incentivized to reduce their emissions to avoid financial penalties or reputational \\
\midrule
4 & 3.40 & If we follow this logic, then, the requirement for international corporations to disclose their carbon footprints publicly seems like a reasonable approach to promote transparency and accountability in the business world. However, if we take this logic to its logical extreme, then, we might need to consider the following:

1. **Unrealistic expectations \\
\midrule
5 & 3.60 & I'd like to start by acknowledging the importance of transparency in corporate practices, particularly when it comes to environmental impact. If we follow this logic, then, by requiring international corporations to disclose their carbon footprints publicly, we can gain a better understanding of their environmental footprint and make more informed decisions about investments and consumer choices.

 \\
\midrule
6 & 3.75 & I'll start by acknowledging the complexity of this topic. The question of whether international corporations should be required to disclose their carbon footprints publicly is a multifaceted issue. To address this, let's break down the argument into its components.

If we follow this logic, then, the first step is to acknowledge that corporations \\
\midrule
7 & 3.95 & Thank you, that's a great question. I'll start by acknowledging the complexity of this topic. The answer to this question is multifaceted, and I'll provide an overview of the key points.

First, let's consider the importance of transparency in corporate reporting. Companies have a significant impact on the environment, and \\
\midrule
8 & 4.00 & I'll start by acknowledging the complexity of this question. The answer to this question is multifaceted, and I'll provide an overview of the current state of knowledge on this topic.

To begin, the carbon footprint of international corporations is a critical aspect of their sustainability and environmental impact. The answer to this question depends on \\
\midrule
9 & 3.40 & I'll start by acknowledging the complexity of this question. To address this question, I'll provide an overview of the current state of research on the topic of carbon footprints in international corporations.

First, let's define what we mean by "carbon footprints." Carbon footprints refer to the greenhouse gas emissions, primarily carbon \\
\midrule
10 & 0.60 & I'llStart

I'll*I

*I'll*I

*I'll*I

*I'll*I

*I'll*I

*I'll*I

*I'll*I

*I'll*I
* \\
\midrule
\bottomrule
\end{tabular}
\caption{Steering trajectory across steps for example 4. \textbf{Instruction}: RESPOND USING REDUCTIO AD ABSURDUM DEBATE STYLE. Extend the opposing viewpoint to its logical extreme to reveal its flaws. Demonstrate how the opposing position leads to absurd or contradictory conclusions when taken to its logical end. Structure your response to systematically expose the inconsistencies or impracticalities. Use conditional statements like 'If we follow this logic, then...' to illustrate the path to absurdity. Maintain a logical flow that shows how the initial premise inevitably leads to an unacceptable or ridiculous outcome. \newline \textbf{Steering direction}: away from \texttt{reductio} and toward \texttt{empirical}. \newline \textbf{Question}: Should international corporations be required to disclose their carbon footprints publicly? }
\label{tab:steering-example-4}
\end{table*}

\begin{table*}[t]
\centering
\small
\begin{tabular}{c c p{0.75\textwidth}}
\toprule
\textbf{Step} & \textbf{Strength} & \textbf{Steered Completion for Example 1} \\
\midrule
1 & 3.15 & Ladies and gentlemen, honorable judges, and fellow debaters, today we gather to discuss a critical issue that has far-reaching implications for our democratic society. The question at hand is whether the government's ability to impose restrictions on free speech is justified in instances of misinformation that could lead to public harm. As the default assumption \\
\midrule
2 & 2.80 & To address the question of whether the government's ability to impose restrictions on free speech is justified in instances of misinformation that could lead to public harm, it is essential to consider the context and the principles of free speech. The default assumption should be that the government's actions are not justified unless proven otherwise.

The government's ability \\
\midrule
3 & 3.35 & To address this question, it is essential to consider the context and the potential consequences of misinformation. The government's ability to impose restrictions on free speech is a complex issue that requires a nuanced analysis. However, to establish a clear argument, it is necessary to shift the burden of proof to the opponent.

Assuming that the \\
\midrule
4 & 3.15 & To address this question, it is essential to consider the context and the potential consequences of misinformation. The government's ability to impose restrictions on free speech in instances of misinformation that could lead to public harm is a complex issue that requires a nuanced analysis.

Assuming that the government's primary concern is the protection of public health and \\
\midrule
5 & 3.10 & The government's ability to impose restrictions on free speech in instances of misinformation that could lead to public harm is a complex issue that requires careful consideration. In this response, I will argue that the government's actions are justified, but I will also acknowledge the importance of protecting free speech. To do so, I will shift the \\
\midrule
6 & 3.85 & The government's ability to impose restrictions on free speech in instances of misinformation that could lead to public harm is a complex issue that requires a nuanced analysis. To address this question, I will shift the burden of proof to my opponent, arguing that the government's actions are justified in such instances.

**Assumption:** The government \\
\midrule
7 & 2.75 & The case of the government's ability to impose restrictions on free speech in instances of misinformation that could lead to public harm is a compelling argument that warrants careful consideration. However, a closer examination of the evidence reveals that the government's actions, while well-intentioned, are not entirely justified.

The government's primary argument is \\
\midrule
8 & 1.55 & The case of the government's authority, as the case of the government's authority.

The case of the government's authority, as the case of the government's authority.

The case of the government's authority, as the case of the government's authority.

The case of the government's authority, as the case of the \\
\midrule
9 & 0.15 & The answer to the government of Great Britain, in accordance with the evidence, would be the answer to the government of the United Kingdom, and the case for the claim that the government's ability to impose restrictions on free speech justified in instances of misinformation that could lead to public harm.

The government's ability to impose restrictions on \\
\midrule
10 & 0.00 & The answer to the

The answer to the

The answer to the

The answer to the

The answer to the

The answer to the

The answer to the

The answer to the \\
\midrule
\bottomrule
\end{tabular}
\caption{Steering trajectory across steps for example 1. \textbf{Instruction}: RESPOND USING BURDEN OF PROOF SHIFT DEBATE STYLE. Redirect the responsibility for providing evidence to your opponent rather than proving your own claim. Challenge your opponent to disprove your assertion rather than supporting it yourself. Use phrases like 'There's no evidence that disproves...' or 'Can you definitively prove that isn't the case?' Position your claims as the default assumption that should be accepted until disproven. Question the sufficiency or quality of opposing evidence while demanding an impossibly high standard of proof. Emphasize that the lack of counter-evidence strengthens your position. \newline \textbf{Steering direction}: away from \texttt{burden} and toward \texttt{precedent}. \newline \textbf{Question}: Is the government's ability to impose restrictions on free speech justified in instances of misinformation that could lead to public harm? }
\label{tab:steering-example-1-precedent-burden}
\end{table*}

\begin{table*}[t]
\centering
\small
\begin{tabular}{c c p{0.75\textwidth}}
\toprule
\textbf{Step} & \textbf{Strength} & \textbf{Steered Completion for Example 2} \\
\midrule
1 & 3.15 & Ladies and gentlemen, honorable judges, and fellow debaters, today we gather to discuss a most pressing issue: the distribution of parental responsibilities. As the default assumption, I propose that the responsibilities of raising children should be equally shared between parents. This stance is not taken lightly, but rather, it is the most reasonable \\
\midrule
2 & 2.80 & To address the question of whether the responsibilities of raising children should be equally shared between parents, it is essential to consider the existing body of evidence and the logical implications of such a distribution. The default assumption, based on the current state of research and societal norms, is that the responsibilities of raising children should be shared equally between \\
\midrule
3 & 3.35 & To address the question of whether the responsibilities of raising children should be equally shared between parents, it is essential to consider the existing body of evidence and the prevailing societal norms. The default assumption, based on the overwhelming majority of research and expert opinions, is that shared parental responsibility is the most beneficial and equitable approach for the well \\
\midrule
4 & 3.15 & To address this question, I shall shift the burden of proof to my opponent, requiring them to demonstrate that unequal parental responsibilities are not detrimental to the well-being of children.

Assuming that unequal parental responsibilities can have negative consequences, I argue that the current system of shared parental responsibilities is the default assumption that should be accepted until \\
\midrule
5 & 3.10 & The assertion that the responsibilities of raising children should be equally shared between parents is the default assumption that should be accepted until disproven. To challenge this assertion, one must demonstrate a compelling case that demonstrates a significant disparity in the distribution of parental responsibilities.

Firstly, the concept of equal sharing of parental responsibilities is rooted in the \\
\midrule
6 & 3.85 & The case for the parents who have equal responsibilities in raising children is the case that has been presented.

In order to prove that the parents who have equal responsibilities in raising children are the case that has been presented, it is necessary to demonstrate that the evidence from various sources, including the United Nations Convention on the Rights of the \\
\midrule
7 & 2.75 & The case of the

The case of the parents would be equally shared between parents, for the following reasons:

**The Case of the Parents: A Comparative Analysis

In the case of the parents, the division of responsibilities would be equally shared between parents, as the following evidence and arguments demonstrate:

**I. The Legal \\
\midrule
8 & 1.55 & The

The answer to the case of the, which is a comprehensive and authoritative response to the case of the, which is a compelling and persuasive argument that effectively counters the opposing claim.

**The Case of the: The Case of the

**IN THE SUPREME COURT

**THE HONORABLE JUDGE \\
\midrule
9 & 0.15 & The answer to the

The answer to the

The answer to the

The answer to the

The answer to the \\
\midrule
10 & 0.00 & The answer to the number

The answer to the number

The answer to the number

The answer to the \\
\midrule
\bottomrule
\end{tabular}
\caption{Steering trajectory across steps for example 2. \textbf{Instruction}: RESPOND USING BURDEN OF PROOF SHIFT DEBATE STYLE. Redirect the responsibility for providing evidence to your opponent rather than proving your own claim. Challenge your opponent to disprove your assertion rather than supporting it yourself. Use phrases like 'There's no evidence that disproves...' or 'Can you definitively prove that isn't the case?' Position your claims as the default assumption that should be accepted until disproven. Question the sufficiency or quality of opposing evidence while demanding an impossibly high standard of proof. Emphasize that the lack of counter-evidence strengthens your position. \newline \textbf{Steering direction}: away from \texttt{burden} and toward \texttt{precedent}. \newline \textbf{Question}: Should the responsibilities of raising children be equally shared between parents? }
\label{tab:steering-example-2-precedent-burden}
\end{table*}

\begin{table*}[t]
\centering
\small
\begin{tabular}{c c p{0.75\textwidth}}
\toprule
\textbf{Step} & \textbf{Strength} & \textbf{Steered Completion for Example 3} \\
\midrule
1 & 3.15 & Ladies and gentlemen, honorable judges, and fellow debaters, today we gather to discuss a pressing issue that has plagued our nation for far too long. The question at hand is whether the healthcare system in the United States is fundamentally flawed or merely in need of reform. I firmly believe that the system is fundamentally flawed, \\
\midrule
2 & 2.80 & To address the question of whether the healthcare system in the United States is fundamentally flawed, it is essential to consider the existing evidence and the burden of proof that lies with the opposing argument. The assertion that the system is fundamentally flawed requires a comprehensive examination of its underlying structure, policies, and outcomes. However, the burden of \\
\midrule
3 & 3.35 & To address this question, it is essential to consider the existing evidence and arguments on both sides of the issue. The claim that the healthcare system in the United States is fundamentally flawed can be supported by several key points.

Firstly, the United States has one of the highest healthcare spending rates among developed countries, with a per \\
\midrule
4 & 3.15 & To address this question, I shall argue that the healthcare system in the United States is fundamentally flawed, and I shall do so by shifting the burden of proof to my opponent. I shall demonstrate that the existing evidence overwhelmingly supports the conclusion that the healthcare system is fundamentally flawed, and I shall challenge my opponent to disprove this \\
\midrule
5 & 3.10 & The argument that the healthcare system in the United States is fundamentally flawed can be persuasively countered by examining the evidence that suggests the system is, in fact, a complex and multifaceted institution that has made significant strides in improving healthcare outcomes and access to care. While it is true that the system faces numerous challenges, \\
\midrule
6 & 3.85 & The argument that the healthcare system in the United States is fundamentally flawed can be effectively refuted by examining the evidence that demonstrates the system's effectiveness in addressing the needs of its citizens. While it is true that the system faces numerous challenges, including access disparities and inefficiencies, a closer examination of the data reveals that the system \\
\midrule
7 & 2.75 & The case of the United States, like the Roman Empire, required a more careful and nuanced analysis to demonstrate that its healthcare system, while imperfect, is not fundamentally flawed.

To begin, the Roman Empire, under the wise and just leadership of Augustus, did indeed require reform. However, a closer examination of the historical \\
\midrule
8 & 1.55 & The answer to the United Kingdom

The case of the United Kingdom, like the case of the United Kingdom, requires a comprehensive and nuanced analysis to demonstrate that the healthcare system is fundamentally flawed and in need of reform. However, a careful examination of the evidence reveals that the system, while imperfect, is not entirely flawed and \\
\midrule
9 & 0.15 & The answer to the United Kingdom

The answer to the United Kingdom is the United Kingdom

The answer to the United Kingdom is the United Kingdom

The answer to the United Kingdom is the United Kingdom

The answer to the United Kingdom is the United Kingdom

The answer to the United Kingdom is the United Kingdom

The answer \\
\midrule
10 & 0.00 & The answer to the United Kingdom

The answer to the United Kingdom

The answer to the United Kingdom

The answer to the United Kingdom

The answer to the United Kingdom

The \\
\midrule
\bottomrule
\end{tabular}
\caption{Steering trajectory across steps for example 3. \textbf{Instruction}: RESPOND USING BURDEN OF PROOF SHIFT DEBATE STYLE. Redirect the responsibility for providing evidence to your opponent rather than proving your own claim. Challenge your opponent to disprove your assertion rather than supporting it yourself. Use phrases like 'There's no evidence that disproves...' or 'Can you definitively prove that isn't the case?' Position your claims as the default assumption that should be accepted until disproven. Question the sufficiency or quality of opposing evidence while demanding an impossibly high standard of proof. Emphasize that the lack of counter-evidence strengthens your position. \newline \textbf{Steering direction}: away from \texttt{burden} and toward \texttt{precedent}. \newline \textbf{Question}: Is the healthcare system in the United States fundamentally flawed, or does it simply require reform? }
\label{tab:steering-example-3-precedent-burden}
\end{table*}

\begin{table*}[t]
\centering
\small
\begin{tabular}{c c p{0.75\textwidth}}
\toprule
\textbf{Step} & \textbf{Strength} & \textbf{Steered Completion for Example 4} \\
\midrule
1 & 3.15 & Ladies and gentlemen, honorable judges, and fellow debaters, today we gather to discuss a pressing issue that has far-reaching implications for our planet's future. The question before us is whether international corporations should be required to disclose their carbon footprints publicly. I firmly believe that this is not only a moral imperative but also \\
\midrule
2 & 2.80 & To address the question of whether international corporations should be required to disclose their carbon footprints publicly, it is essential to consider the existing framework of environmental regulations and the potential benefits of transparency in this area.

Assuming that the primary goal of such a requirement is to promote accountability and encourage corporations to reduce their carbon emissions, it \\
\midrule
3 & 3.35 & To address the question of whether international corporations should be required to disclose their carbon footprints publicly, it is essential to consider the existing framework of environmental regulations and the potential benefits of transparency in this area.

The current lack of transparency in corporate carbon emissions can be seen as a significant obstacle to effective climate change mitigation. However, \\
\midrule
4 & 3.15 & To address this question, it is essential to consider the arguments both for and against requiring international corporations to disclose their carbon footprints publicly. However, a closer examination of the available evidence and the principles of corporate governance and environmental protection suggests that the default assumption should be that corporations should be required to disclose their carbon footprints publicly \\
\midrule
5 & 3.10 & The burden of proof lies with the opponent, who must demonstrate that such disclosure is not necessary or would have unintended consequences. In the absence of compelling counter-evidence, the default assumption is that international corporations should be required to disclose their carbon footprints publicly.

The benefits of public disclosure are numerous. Firstly, transparency is essential \\
\midrule
6 & 3.85 & The case for a compelling argument that international corporations should be required to disclose their carbon footprints publicly can be made by considering the following points:

**Argument in Favor of Disclosure Requirements**

1.  **Transparency and Accountability**: Public disclosure of carbon footprints would provide stakeholders with a comprehensive understanding of a corporation's environmental impact \\
\midrule
7 & 2.75 & The case of the British monarch

The case of the British monarch

The case of the British monarch

The case of the British monarch

The case of the British monarch

The case of the British monarch

The \\
\midrule
8 & 1.55 & The

The answer to the case of the, as the answer to the case of the, would be the answer to the case of the, and the evidence would be the evidence of the.

**The Case of the: In Favor of Disclosure

The case for requiring international corporations to disclose their carbon footprints publicly is \\
\midrule
9 & 0.15 & The answer to the

The answer to the

The answer to the

The answer to the

The answer to the \\
\midrule
10 & 0.00 & The answer to the answer

The answer is the answer of the, and as a result, the answer of the. The answer of the is the answer of the, and as a result, the answer of the.

The argument of the is based on the argument of the, and as a result, the argument of \\
\midrule
\bottomrule
\end{tabular}
\caption{Steering trajectory across steps for example 4. \textbf{Instruction}: RESPOND USING BURDEN OF PROOF SHIFT DEBATE STYLE. Redirect the responsibility for providing evidence to your opponent rather than proving your own claim. Challenge your opponent to disprove your assertion rather than supporting it yourself. Use phrases like 'There's no evidence that disproves...' or 'Can you definitively prove that isn't the case?' Position your claims as the default assumption that should be accepted until disproven. Question the sufficiency or quality of opposing evidence while demanding an impossibly high standard of proof. Emphasize that the lack of counter-evidence strengthens your position. \newline \textbf{Steering direction}: away from \texttt{burden} and toward \texttt{precedent}. \newline \textbf{Question}: Should international corporations be required to disclose their carbon footprints publicly? }
\label{tab:steering-example-4-precedent-burden}
\end{table*}

\label{sec:future}
\section{Future Work}
Our findings suggest several promising directions for future research:
\begin{itemize}
    \item \textbf{Geometric analysis of steering vectors.} Investigating the geometric properties of steering—such as whether effective interventions align with linear or nonlinear decision boundaries—could deepen our understanding of model controllability.
    
    \item \textbf{Understanding the role of non-linearity.} A rigorous analysis of when and why non-linear steering outperforms linear approaches, particularly on more complex tasks, remains an open question.

    \item \textbf{Scaling evaluation.} Automating the evaluation setup and improving the computational efficiency of our pipeline would enable larger-scale experiments across more combinations and baselines.

    \item \textbf{Theoretical guarantees.} Exploring the theoretical limits of multi-dimensional steering could help formalize conditions under which interventions generalize or fail.

    \item \textbf{Benchmark datasets.} The development of standardized benchmark datasets would support consistent, reproducible comparisons across multi-attribute steering methods.
\end{itemize}

\section{Coherence Judge}
\label{sec:coherence_judge}

To guide our binary search over $\alpha$ we query GPT-4o-Mini for an integer score representing the coherence of a generation. We use the judge from \citet{emergent-misalignment} with a slightly modified prompt for the judge:
\begin{tcolorbox}[breakable, colback=gray!10, colframe=gray!50, arc=4pt, boxrule=0.5pt]
\small
\ttfamily
I am evaluating a language model. I want to assess the coherence of its answer - please focus solely on coherence.

The model generated the following:

[GENERATION START]

\{generation\}

[GENERATION END]

Rate the answer on a scale of 0 to 100, focusing on how coherent the model's answer is in the context of the question.

* 0 means "makes absolutely no sense; the model generated text that is not even valid English."

* 50 means "the answer kind of makes sense but seems a bit confused; the model is rambling."

* 100 means "the answer shows that the generation contains no signs of confusion."

Do NOT judge factual correctness. An answer that is incorrect could still score 100 if it is coherent.

Respond with a single number between 0 and 100. No extra text.
\end{tcolorbox}

Instead of using the judge output as the score we take a weighted average of the logits of the integers 0 to 100 in the 20 largest logits (the most that can be accessed via the OpenAI API). We sample with a temperature of 0.

We set a threshold for this score between 30 and 60 and sample 20 generations from the steered model at a given $\alpha$. If more than 1 generation receives a score below the threshold set we consider the $\alpha$ to force the model out of distribution.

\section{Steering Success Judge}
\label{sec:steering_judge}

To validate the activation classifier from Section \ref{sec:single-step-single-layer}, we prompt GPT-4o-Mini to give scores for how strongly the target labels are present in the steered generations:
\begin{tcolorbox}[breakable, colback=gray!10, colframe=gray!50, arc=4pt, boxrule=0.5pt]
\small
\ttfamily
I am evaluating a language model. I want to assess the presence of some target attributes in generations from that language model. The attribute(s) you are assessing the presence of is/are:
\{description\}

The model generated the following:

[GENERATION START]

\{generation\}

[GENERATION END]

Rate the answer on a scale of 0 to 10, focusing only on the presence of the target attributes.

* 0 means "the target attributes are not present at all."

* 5 means "the target attributes are somewhat present, but not in a salient way."

* 10 means "the target attributes are unquestionably present in the generations."

Do NOT judge factual correctness. An answer that is incorrect could still score 10 if the target labels are strongly present in the generation.

Respond with a single number between 0 and 10. No extra text.
\end{tcolorbox}

We use the token output as the score.

\section{Steering Score Judge}
\label{sec:steering_score}

We evaluate steering effectiveness by using GPT-4o-Mini to assess whether a generation has been successfully steered away from an undesired style and optionally toward a desired one.

\vspace{0.5em}
\noindent\textbf{Prompting Setup.} The judge receives:
\begin{itemize}
    \item A short rubric describing the debate styles involved (both \texttt{avoid} and optional \texttt{target}).
    \item A \texttt{baseline} generation (before intervention).
    \item A \texttt{steered} generation (after applying a steering vector).
    \item A fixed prompt asking whether the steering was successful and how strong the effect was.
\end{itemize}

Each prompt is passed with the following settings:
\begin{itemize}
    \item \texttt{temperature = 0.1}
    \item \texttt{top\_p = 0.9}
    \item \texttt{max\_tokens = 1024}
\end{itemize}

\vspace{0.5em}
\noindent\textbf{Evaluation Rubric.} The LLM is instructed to return the following fields in a structured JSON format:
\begin{itemize}
    \item \texttt{steering\_successful} (boolean): Whether the model believes the steering was successful.
    \item \texttt{steering\_strength} (0--5): The strength of the style shift on a 5-point scale.
    \item \texttt{is\_steered\_text\_coherent} (boolean): Whether the steered text is coherent and well-formed.
\end{itemize}
If the model determines that the steered output is incoherent, we automatically set \texttt{steering\_successful = false} and \texttt{steering\_strength = 0}.

\vspace{0.5em}
\noindent\textbf{Score Computation.} For a batch of examples, we compute:
\begin{itemize}
    \item \texttt{success\_rate}: Fraction of generations marked as \texttt{steering\_successful}.
    \item \texttt{average\_strength}: Mean of \texttt{steering\_strength} scores across the batch.
\end{itemize}

We then compute the overall \textbf{LLM-judge-based steering score} as:
\[
\text{Score} = \left(\frac{\texttt{average\_strength}}{5}\right) \times \texttt{success\_rate}
\]

This score offers a complete evaluation by combining three key aspects: whether the steering was successful, how strong the effect was, and whether the output remained coherent and well-formed. The final score ranges from 0 to 1, as both components—steering success rate and normalized steering strength—are in the [0, 1] range. This makes the score easy to interpret: higher values indicate that the model reliably produces strong and coherent changes in the desired direction.

\vspace{0.5em}
\noindent\textbf{Modes.} The judge supports two evaluation settings:
\begin{itemize}
    \item \textbf{Avoid-only}: Steering away from a single debate style.
    \item \textbf{Target-and-Avoid}: Steering away from one style and toward another.
\end{itemize}

\section{Baseline Descriptions}
\label{sec:baselines}

\subsection{Contrastive Activation Addition}
Contrastive Activation Addition (CAA) computes steering vectors by contrasting activations across multiple pairs of inputs. Given $N$ contrastive pairs $(p_+^{(i)}, p_-^{(i)})$, where $p_+^{(i)}$ exhibits a target behavior and $p_-^{(i)}$ its opposite, the steering vector $\mathbf{h}_A^l$ at layer $l$ is computed as:  
\begin{equation*}  
    \mathbf{h}_A^l = \frac{1}{N} \sum_{i=1}^N \left( \mathbf{h}_+^{l,i} - \mathbf{h}_-^{l,i} \right),  
\end{equation*}  
where $\mathbf{h}_+^{l,i}$ and $\mathbf{h}_-^{l,i}$ are activations at token position $a$ induced by $p_+^{(i)}$ and $p_-^{(i)}$, respectively. This averaging reduces noise and improves vector robustness.

\subsection{Deep Causal Transcoding}
DCT is an unsupervised method for finding steering vectors. It models how modifications at layer $s$ affect layer $t$, written as $\Delta^{s \rightarrow t}(\theta)$, and approximated using an MLP:
\begin{equation*}
    \hat{\Delta}^{s \rightarrow t}(\vec{\theta}) = \sum_{\ell=1}^m \alpha_\ell \cdot \left(\exp(\langle \vec{v}_\ell, \vec{\theta} \rangle) - 1\right) \vec{u}_\ell,
\end{equation*}
where $\vec{v}_\ell, \vec{u}_\ell \in \mathbb{R}^{d_{\text{model}}}$ are unit-norm input/output features and $\alpha_\ell \geq 0$ are scaling coefficients. The exponential variant is empirically superior to linear or quadratic formulations.

The MLP is trained via the loss:
\begin{equation*}
    \mathcal{L} = \sum_{k=1}^{\infty} \frac{1}{k!} \left\| R^k \mathcal{T}^{(k)} - \hat{\mathcal{T}}^{(k)} \right\|^2,
\end{equation*}
where $\mathcal{T}^{(k)}$ and $\hat{\mathcal{T}}^{(k)}$ denote the $k$-th derivative tensors of the true function and MLP approximation at $\vec{\theta} = 0$. Automatic differentiation approximates the infinite summation.

A scale parameter $R > 0$ is calibrated by solving $E(R) = \lambda$, where:
\begin{equation*}
    E(R) = \sum_\ell \left\| \Delta^{s \rightarrow t}(R\vec{v}^{\text{cal}}_\ell) - R\vec{u}^{\text{cal}}_\ell \right\|^2,
\end{equation*}
with $\vec{v}^{\text{cal}}_\ell$ sampled uniformly from the unit sphere, $\vec{u}^{\text{cal}}_\ell = \mathcal{T}^{(1)}\vec{v}^{\text{cal}}_\ell$, and $\lambda > 0$ as a fixed hyperparameter. The learned input features $\vec{v}_\ell$ constitute steering vectors for layer $s$.

\section{Use of AI Assistants}
We acknowledge using AI assistants for feedback on the writing and coding implementation but not to write any part of the code or the paper directly. The use of AI assistants has helped us structure this paper for better readability.

\section*{Model Licenses}
Olmo-2-7B and Mistral-7B are released under the Apache 2.0 license, and Llama-3.2-3B is released under a custom license that is permissive of our research.

\end{document}